\documentclass[twoside,11pt]{article}

\usepackage{jmlr2e}
\usepackage{amsmath}
\usepackage{algorithm}
\usepackage{algorithmic}
\usepackage{subfigure}
\usepackage{graphicx}
\usepackage{bbm}
\usepackage{tikz}
\usetikzlibrary{arrows}
\usetikzlibrary{graphs}
\usepackage{url}

\newcommand{\mb}{\mathbf}
\newcommand{\mc}{\mathcal}
\newcommand{\bs}{\boldsymbol}

\newcommand{\gcn}{\textsc{GCN}}
\newcommand{\gat}{\textsc{GAT}}
\newcommand{\sage}{\textsc{GraphSage}}
\newcommand{\gdu}{\textsc{GDU}}
\newcommand{\dif}{\textsc{DifNN}}
\newcommand{\isonn}{\textsc{IsoNN}}
\newcommand{\sdbn}{\textsc{SDBN}}
\newcommand{\lfer}{\textsc{LF\&ER}}
\newcommand{\segen}{\textsc{seGEN}}
\newcommand{\gnl}{\textsc{GNL}}
\newcommand{\lasso}{\textsc{Lasso}}

\ShortHeadings{IFM LAB TUTORIAL SERIES \# 7, COPYRIGHT \copyright IFM LAB}{JIAWEI ZHANG, IFM LAB DIRECTOR}
\firstpageno{1}

\begin{document}

\title{Graph Neural Networks for Small Graph and Giant Network Representation Learning: An Overview}

\author{\name Jiawei Zhang \email jiawei@ifmlab.org \\
	\addr{Founder and Director}\\
       {Information Fusion and Mining Laboratory}\\
       (First Version: July 2019; Revision: July 2019.)}

\maketitle

\begin{abstract}

Graph neural networks denote a group of neural network models introduced for the representation learning tasks on graph data specifically. Graph neural networks have been demonstrated to be effective for capturing network structure information, and the learned representations can achieve the state-of-the-art performance on node and graph classification tasks. Besides the different application scenarios, the architectures of graph neural network models also depend on the studied graph types a lot. Graph data studied in research can be generally categorized into two main types, i.e., {small graphs} \textit{vs.} {giant networks}, which differ from each other a lot in the \textit{size}, \textit{instance number} and \textit{label annotation}. Several different types of graph neural network models have been introduced for learning the representations from such different types of graphs already. In this paper, for these two different types of graph data, we will introduce the graph neural networks introduced in recent years. To be more specific, the graph neural networks introduced in this paper include {\isonn} \cite{isonn}, {\sdbn} \cite{sdbn}, {\lfer} \cite{lfer}, {\gcn} \cite{gcn}, {\gat} \cite{gat}, {\dif} \cite{dif}, {\gnl} \cite{gnl}, {\sage} \cite{sage} and {\segen} \cite{segen}. Among these graph neural network models, {\isonn}, {\sdbn} and {\lfer} are initially proposed for small graphs and the remaining ones are initially proposed for giant networks instead. The readers are also suggested to refer to these papers for detailed information when reading this tutorial paper.

\end{abstract}


\tableofcontents

\section{Introduction}

In the era of big data, graph provides a generalized representation of many different types of inter-connected data collected from various disciplines. Besides the unique attributes possessed by individual nodes, the extensive connections among the nodes can convey very complex yet important information. Graph data are very difficult to deal with because of their \textit{various shapes} (e.g., small brain graphs \textit{vs.} giant online social networks), \textit{complex structures} (containing various kinds of nodes and extensive connections) and \textit{diverse attributes} (attached to the nodes and links). Great challenges exist in handling the graph data with traditional machine learning algorithms directly, which usually take feature vectors as the input. Viewed in such a perspective, learning the feature vector representations of graph data will be an important problem.

The graph data studied in research can be generally categorized into two main types, i.e., {small graphs} \textit{vs.} {giant networks}, which differ from each other a lot in the \textit{size}, \textit{instance number} and \textit{label annotation}. 
\begin{itemize}
\item The small graphs we study are generally of a much smaller size, but with a large number of instances, and each graph instance is annotated with certain labels. The representative examples include the human brain graph, molecular graph, and real-estate community graph, whose nodes (usually only in hundreds) represent the brain regions, atoms and POIs, respectively.

\item On the contrary, giant networks in research usually involve a large number of nodes/links, but with only one single network instance, and individual nodes are labeled instead of the network. Examples of giant networks include {social network} (e.g., Facebook), {eCommerce network} (e.g., Amazon) and {bibliographic network} (e.g., DBLP), which all contain millions even billions of nodes.
\end{itemize}

Due to these property distinctions, the representation learning algorithms proposed for small graphs and giant networks are very different. To solve the small graph oriented problems, the existing graph neural networks focus on learning a representation of the whole graph (not the individual nodes) based on the graph structure and attribute information. Several different graph neural network models have been introduced already, including {\isonn} (Isomorphic Neural Network) \cite{isonn}, {\sdbn} (Structural Deep Brain Network) \cite{sdbn} and {\lfer} (Deep Autoencoder based Latent Feature Extraction) \cite{lfer}. These models are proposed for different small graph oriented application scenarios, covering both brain graphs and community POI graphs, which can also be applied to other application settings with minor extensions.

Meanwhile, for the giant network studies, in recent years, many research works propose to apply graph neural networks to learn their low-dimensional feature representations, where each node is represented as a feature vector. With these learned node representations, the graph neural network model can directly infer the potential labels of the nodes/links. To achieve such objectives, several different type of graph neural network models have been introduced, including {\gcn} (Graph Convolutional Network) \cite{gcn}, {\gat} (Graph Attention Network) \cite{gat}, {\dif} (Deep Diffusive Neural Network) \cite{dif}, {\gnl} (Graph Neural Lasso), {\sage} (Graph Sample and Aggregate) \cite{sage} and {\segen} (Sample and Ensemble Genetic Evolutionary Network) \cite{segen}.

In this paper, we will introduce the aforementioned graph neural networks proposed for small graphs and giant networks, respectively. This tutorial paper will be updated accordingly as we observe the latest developments on this topic.


\section{Graph Neural Networks for Small Graphs}\label{sec:small_graph}

In this section, we will introduce the graph neural networks proposed for the representation learning tasks on small graphs. Formally, we can represent the set of small graphs to be studied in this section as $\mc{G} = {(G_1, \mb{y}_1), (G_2, \mb{y}_2), \cdots, (G_n, \mb{y}_n)}$, where $G_i = (\mc{V}_i, \mc{E}_i)$ denotes a small graph instance and $\mb{y}_i \in \mathbbm{R}^{d_y}$ denotes its label vector. Given a graph $G_i \in \mc{G}$, we can denote its network size as the number of involved nodes, i.e., $|\mc{V}_i|$. Normally, the small graphs to be studied in set $\mc{G}$ are of the same size. Meanwhile, depending on the application scenarios, the objective labels of the graph instances can be binary-class/multi-class vectors. The small graph oriented graph neural networks aim at learning a mapping, i.e., $f: \mc{G} \to \mathbbm{R}^{d_h}$, to project the graph instances to their feature vector representations, which will be further utilized to infer their corresponding labels. Specifically, the graph neural network models to be introduced in this section include {\isonn} \cite{isonn}, {\sdbn} \cite{sdbn} and {\lfer} \cite{lfer}. The readers are also suggested to refer to these papers for detailed information when reading this tutorial paper.


\subsection{{\isonn}: Isomorphic Neural Network}

\begin{figure}[t]
    \centering
    \hspace*{-3em}
    \includegraphics[width=1.2\textwidth]{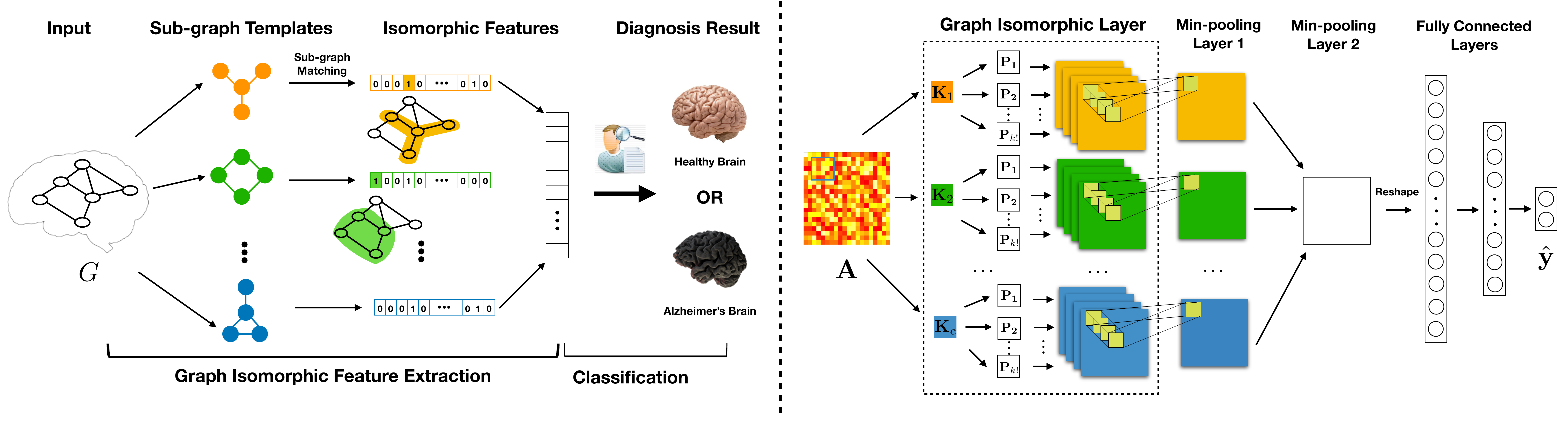}
    \caption{{\isonn} Model Architecture \cite{isonn}. (The left plot provides the outline of {\isonn}, including the \textit{graph isomorphic feature extraction} and \textit{classification} components. The right plot illustrates the detailed architecture of the proposed model, where the \textit{graph isomorphic features} are extracted with the \textit{graph isomorphic layer} and two \textit{min-pooling layers}, and the graphs are further classified with three \textit{fully-connected layers}.)} 
    \label{fig:model}
\end{figure} 

Graph isomorphic neural network ({\isonn}) proposed in \cite{isonn} recently aims at extracting meaningful sub-graph patterns from the input graph for representation learning. Sub-graph mining techniques have been demonstrated to be effective for feature extraction in the existing works. Instead of designing the sub-graph templates manually, {\isonn} proposes to integrate the sub-graph based feature extraction approaches into the neural network framework for automatic feature representation learning. As illustrated in Figure~\ref{fig:model}, {\isonn} includes two main components: \textit{graph Isomorphic feature extraction} component and \textit{classification} component. {\isonn} can be in a deeper architecture by involving multiple graph isomorphic feature extraction components so that the model will learn more complex sub-graph patterns.

The graph isomorphic feature extraction component in {\isonn} targets at the automatic sub-graph pattern learning and brain graph feature extraction with the following three layers:\textit{graph isomorphic layer}, \textit{min-pooling layer 1} and \textit{min-pooling layer 2}, which will be introduced as follows, respectively. Meanwhile, the classification component used in {\isonn} involves several fully connected layers, which project the learned isomorphic features to the corresponding graph labels.

\subsubsection{Graph Isomorphic Layer}

In {\isonn}, the sub-graph based feature extraction process is achieved by a novel \textit{graph isomorphic layer}. Formally, given a brain graph $G = (\mathcal{V}, \mathcal{E})$, its adjacency matrix can be represented as $\mb{A} \in \mathbb{R}^{|\mathcal{V}| \times |\mathcal{V}| }$. In order to find the existence of specific sub-graph patterns in the input graph, {\isonn} matches the input graph with a set of sub-graph templates. Instead of defining these sub-graph templates manually as the existing works, each template is denoted as a kernel variable $\mb{K}_i \in \mathbb{R}^{k \times k}, \forall i \in \{1, 2, \cdots, c\}$ and {\isonn} will learn these kernel variables automatically. Here, $k$ denotes the node number in the templates and $c$ is the channel number. Meanwhile, to match one template (i.e., the kernel variable matrix $\mb{K}_i$) with regions in the input graph (i.e., sub-matrices in $\mb{A}$), {\isonn} uses a set of permutation matrices, which map both rows and columns of the kernel matrix to the sub-matrices in $\mb{A}$ effectively. The permutation matrix can be represented as $\mb{P} \in \{0, 1\}^{k \times k}$ that shares the same dimensions with the kernel matrix. Given a kernel variable matrix $\mb{K}_i$ and a regional sub-matrix $\mb{M}_{(s,t)} \in \mathbb{R}^{k \times k}$ in $\mb{A}$ (where $\mb{M}_{(s,t)}(1:k,1:k) = \mb{A}(s:s+k-1,t:t+k-1)$ and index pair $s,t \in \{1, 2, \cdots, (|\mathcal{V}|-k+1)\}$), there may exist $k!$ different such permutation matrices and the optimal one can be denoted as $\mb{P}^*$:
\begin{equation}
\mb{P}^* = \arg \min_{\mb{P} \in \mathcal{P}} \left\| \mb{P}\mb{K}_i\mb{P}^\top - \mb{M}_{(s,t)} \right\|_F^2,
\end{equation}
where $\mathcal{P} = \{\mb{P}_1, \mb{P}_2, \cdots, \mb{P}_{k!}\}$ covers all the potential permutation matrices. The F-norm term measures the mapping loss, which is also used as the graph isomorphic feature in {\isonn}. Formally, the isomorphic feature extracted based on the kernel $\mb{K}_i$ for the regional sub-matrix $\mb{M}_{(s,t)}$ in $\mb{A}$ can be represented as
\begin{equation}
\begin{aligned}
 z_{i, (s,t)}  &= \left\| \mb{P}^*\mb{K}_i (\mb{P}^*)^\top - \mb{M}_{(s,t)} \right\|_F^2 \\
 &= \min \left\{ \left\| \mb{P}\mb{K}_i\mb{P}^\top -\mb{M}_{(s,t)} \right\|_F^2 \right\}_{\mb{P} \in \mathcal{P}} \\
 &= \min( \bar{\mb{z}}_{i,(s,t)} (1:k!) ),
\end{aligned}
\end{equation}
where vector $\bar{\mb{z}}_{i, (s,t)} \in \mathbb{R}^{k!}$ with $\bar{\mb{z}}_{i, (s,t)}(j) = \left\| \mb{P}_j \mb{K}_i \mb{P}_j^\top - \mb{M}_{(s,t)} \right\|_F^2, \forall j \in \{1, 2, \cdots, k!\}$ denoting the features computed on the permutation matrix $\mb{P}_j \in \mc{P}$. Furthermore, by shifting the kernel matrix $\mb{K}_i$ on regional sub-matrices in $\mb{A}$, the isomorphic features extracted by {\isonn} from the input graph can be denoted as a 3-way tensor ${\mathcal{Z}}_i \in \mathbb{R}^{k! \times (|\mathcal{V}|-k+1) \times (|\mathcal{V}|-k+1)}$, where ${\mathcal{Z}}_i(1:k!, s, t) = \bar{\mb{z}}_{i, (s,t)}(1:k!)$.

\subsubsection{Min-pooling Layers}

\noindent $\bullet$ \textbf{Min-pooling Layer 1}: As indicated by the Figure~\ref{fig:model}, {\isonn} computes the final isomorphic features with the optimal permutation matrix for the kernel $\mb{K}_i$ via two steps: (1) computing all the potential isomorphic features via different permutation matrices with the graph isomorphic layer, and (2) identifying the optimal features with the min-pooling layer 1 and layer 2. Formally, given the tensor ${\mathcal{Z}}_i$ computed by $\mb{K}_i$ in the graph isomorphic layer, {\isonn} will identify the optimal permutation matrices via the min-pooling layer 1. From tensor ${\mathcal{Z}}_i$, the features computed with the optimal permutation matrices can be denoted as $\mb{Z}_i$, where
\begin{equation}
\mb{Z}_i(s,t) = \min \{ {\mathcal{Z}}_i(1:k!, s ,t )\}, \forall s,t \in \{1, 2, \cdots, (|\mathcal{V}|-k+1)\}.
\end{equation}
The min-pooling layer 1 learns the optimal feature matrix $\mb{Z}_i$ for kernel $\mb{K}_i$ along the first dimension of tensor $\mc{Z}_i$, which are computed by the optimal permutation matrices. In a similar way, for the remaining kernels, their optimal graph isomorphic features can be obtained and denoted as matrices $\mb{Z}_1$, $\mb{Z}_2$, $\cdots$, $\mb{Z}_c$, respectively.

\noindent $\bullet$ \textbf{Min-pooling Layer 2}: For the same region in the input graph, different kernels can be applied to match the regional sub-matrix. Inspired by this, {\isonn} incorporates the min-pooling layer 2, so that the model can find the best kernels that match the regions in $\mb{A}$. With inputs $\mb{Z}_1, \mb{Z}_2, \cdots, \mb{Z}_c$, the min-pooling layer 2 in {\isonn} can identify the optimal features across all the kernels, which can be denoted as matrix $\mb{Q}$ with
\begin{equation}
\mb{Q}(s,t) = min\{\mb{Z}_1(s,t),  \mb{Z}_2(s,t), \cdots, \mb{Z}_c(s,t) \}, \forall s, t \in \{1, 2, \cdots, (|\mathcal{V}|-k+1)\}.
\end{equation}
Entry $\mb{Q}(s, t)$ denotes the graph isomorphic feature computed by the best sub-graph kernel on the regional matrix $\mb{M}_{(s,t)}$ in $\mb{A}$. Thus, via min-pooling layer 2, let $\mb{Q}$ be the final isomorphic feature matrix, which preserves the best sub-graph patterns contributing to the classification result. In addition, min-pooling layer 2 also effectively shrinks the feature length and greatly reduces the number of variables to be learned in the following classification component. 

\subsubsection{Classification Component}

Given a brain graph instance $G_g \in \mc{B}$ ($\mc{B} \subset \mc{T}$ denotes the training batch), its extracted isomorphic feature matrix can be denoted as $\mb{Q}_g$. By feeding its flat vectorized representation vector $\mb{q}_g =vec(\mb{Q}_g)$ as the input into the classification component (with three fully-connected layers), the predicted label vector by {\isonn} on the instance can be represented as $\hat{\mb{y}}_g$. Several frequently used loss functions, e.g., cross-entropy, can be used to measure the introduced loss between $\hat{\mb{y}}_g$ and the ground-truth label vector $\mb{y}_g$. Formally, the fully-connected (FC) layers and the loss function used in {\isonn} can be represented as follows:
\begin{equation}\mbox{FC Layers: }
\begin{cases}
{\mb{d}_1} ~ = &\sigma(\mb{W}_1 \mb{q}_g + \mb{b}_1), \\
{\mb{d}_2} ~ = &\sigma(\mb{W}_2\mb{d}_1 + \mb{b}_2), \\
\hat{\mb{y}}_g ~~~= & softmax(\mb{W}_3 \mb{d}_2 + \mb{b} _3);
\end{cases}
\end{equation}
and
\begin{equation}
\mbox{ Loss Function: }\ell(\boldsymbol{\Theta}) =  - \sum_{g \in \mathcal{B}} \sum_{j} \mb{y}_g(j) \log \hat{\mb{y}}_g(j),
\end{equation}
where $\mb{W}_i$ and $\mb{b}_i$ are the weight and biase in $i_{th}$ layer, $\sigma(\cdot)$ denotes the sigmoid activation function and $softmax(\cdot)$ is the softmax function for output normalization. Variables $\bs{\Theta} = (\{\mb{K}\}_{i=1}^k, \{\mb{W}_i, \mb{b}_i\}_{i=1}^3)$ (including the kernel matrices and weight/bias terms) involved in the model can be effectively learned with the error back propagation algorithm by minimizing the above loss function. For more information about {\isonn}, the readers are suggested to refer to \cite{isonn} for detailed descriptions.



\subsection{{\sdbn}: Structural Deep Brain Network}

Structural Deep Brain Network ({\sdbn}) initially proposed \cite{sdbn} applies the deep convolutional neural network to the brain graph mining problem, which can be viewed as an integration of convolutional neural network and autoencoder. Similar to {\isonn}, {\sdbn} also observes the order-less property with the brain graph data, and introduce a graph reordering approach to resolve the problem. 

\begin{figure}[t]
    \centering
    \includegraphics[width=1.0\textwidth]{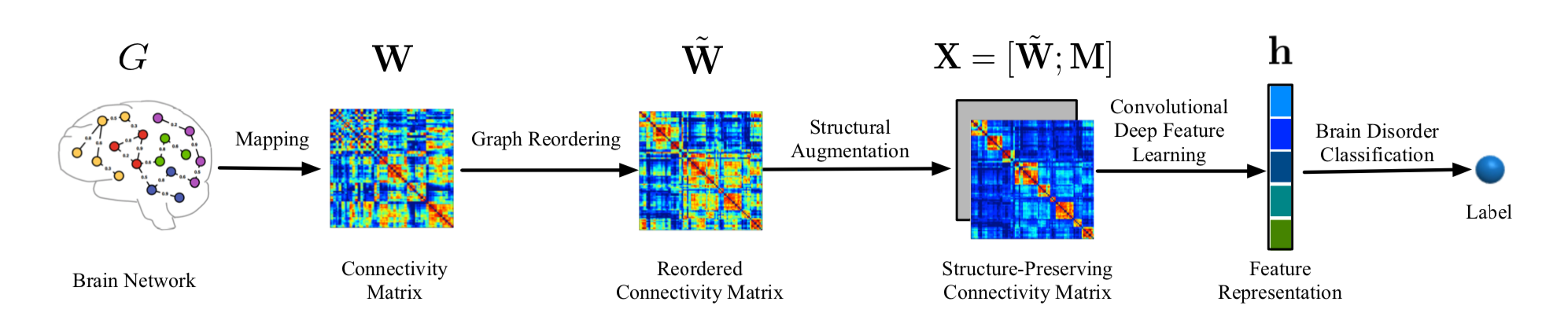}
    \caption{{\sdbn} Model Architecture \cite{sdbn}.} 
    \label{fig:sdbn_model}
\end{figure} 

As illustrated in Figure~\ref{fig:sdbn_model}, besides the necessary graph data processing and representation, {\sdbn} involves three main steps to learn the final representations and labels fo the graph instances, i.e., \textit{graph reordering}, \textit{structural augmentation} and \textit{convolutional feature learning}, which will be introduced as follows, respectively.


\subsubsection{Graph Reordering}

Given the graph set $\mc{G} = \{G_1, G_2, \cdots, G_n\}$, the goal of graph reordering is to find a node labeling $\ell_n$ such that for any two graphs $G_i, G_j \in \mc{G}$ randomly draw from $\mc{G}$, the expected differences between the distance of the graph connectivity adjacency matrices based on $\ell_n$ and the distance of the graphs in the graph space is minimized. Formally, for each graph instance $G_i \in \mc{G}$, its connectivity adjacency matrix can be denoted as $\mb{A}_i$. Let $d_{\mc{A}}$ and $d_{\mc{G}}$ denote the distance metrics on the adjacency matrix domain $\mc{A}$ and graph domain $\mc{G}$ respectively, the graph reordering problem can be formulated as the following optimization problem:
\begin{equation}
\arg \min_{\ell_n} \mathbbm{E}_{G_i, G_j \in \mc{G}} \left( \left\| d_{\mc{A}} (\mb{A}_i, \mb{A}_j) - d_{\mc{G}}(G_i, G_j) \right\| \right).
\end{equation}

Graph reordering is a combinatorial optimization problem, which has also be demonstrated to be NP-hard and is computationally infeasible to address in polynomial time. {\sdbn} proposes to apply the spectral clustering to help reorder the nodes and brain graph connectivity instead.  Formally, based on the brain graph adjacency matrix $\mb{A}_i$ of $G_i$, its corresponding Laplacian matrix can be represented as $\mb{L}_i$. The spectral clustering algorithm aims at partitioning the brain graph $G_i$ into $K$ modules, where the node-module belonging relationships are denoted by matrix $\mb{F} \in \mathbbm{R}^{|\mc{V}| \times K}$. The optimal $\mb{F}$ can be effectively learned with the following objective function:
\begin{equation}
\begin{aligned}
&\min_{\mb{F}} tr\left( \mb{F}^\top \mb{L}_i \mb{F} \right)
&s.t. \mb{F}^\top \mb{F} = \mb{I},
\end{aligned}
\end{equation}
where $\mb{I} \in \{0, 1\}^{K \times K}$ denotes an identity matrix and the constraint is added to ensure one node is assigned to one module only. From the learned optimal $\mb{F}$, {\sdbn} can assign the nodes in graph $G_i$ to their modules $\mc{M} = \{M_1, M_2, \cdots, M_K\}$, where $\mc{V} = \bigcup_{i=1}^K M_i$ and $M_i \cap M_j = \emptyset, \forall i \neq j$ and $i, j \in \{1, 2, \cdots, K\}$. Such learned modules $\mc{M}$ can help reorder the nodes in the graph into relatively compact regions, and the graph connectivity adjacency matrix $\mb{A}_i$ after reordering can be denoted as $\tilde{\mb{A}}_i$. Similar operations will be performed on the other graph instances in the set $\mc{G}$.


\subsubsection{Structural Augmentation}

Prior to feeding the reordered graph adjacency matrix to the deep convolutional neural network for representation learning, {\sdbn} proposes to augment the network structure by both refining the connectivity adjacency matrix and creating an additional module identification channel. 
\begin{itemize}
\item \textbf{Reordered Adjacency Matrix Refinement}: Formally, for graph $G_i \in \mc{G}$, based on its reordered adjacency matrix $\tilde{\mb{A}}_i$ obtained from the previous step, {\sdbn} proposes to refine its entry values with the following equation:
\begin{equation}
\tilde{\mb{A}}_i(p, q)  =
\begin{cases}
1 & \mbox{ for } v_p, v_q \in M_k, \exists M_k \in \mc{M};\\
\epsilon & \mbox{ otherwise}.
\end{cases}
\end{equation}
In the equation, term $\epsilon$ denotes a small constant.

\item \textbf{Module Identification Channel Creation}: From the reordered adjacency matrix $\tilde{\mb{A}}_i$ for graph $G_i \in \mc{G}$, the learned module identity information is actually not preserved. To effectively incorporate such information in the model, {\sdbn} proposes to create one more channel $\mb{M}_i \in \mathbbm{R}^{|\mc{V}| \times |\mc{V}|}$ for graph $G_i$, whose entry values can be denoted as follows:
\begin{equation}
\mb{M}_i(p, q)  =
\begin{cases}
k & \mbox{ for } v_p, v_q \in M_k, \exists M_k \in \mc{M};\\
0 & \mbox{ otherwise}.
\end{cases}
\end{equation}
\end{itemize}
Formally, based on the above operations, the inputs for the representation learning component on graph $G_i$ will be $\mb{X}_i = \left[ \tilde{\mb{A}}_i; \mb{M}_i \right]$, which encodes much more information and can help learn more useful representations.


\subsubsection{Learning of the {\sdbn} Model}

\begin{figure}[t]
    \centering
    \includegraphics[width=0.8\textwidth]{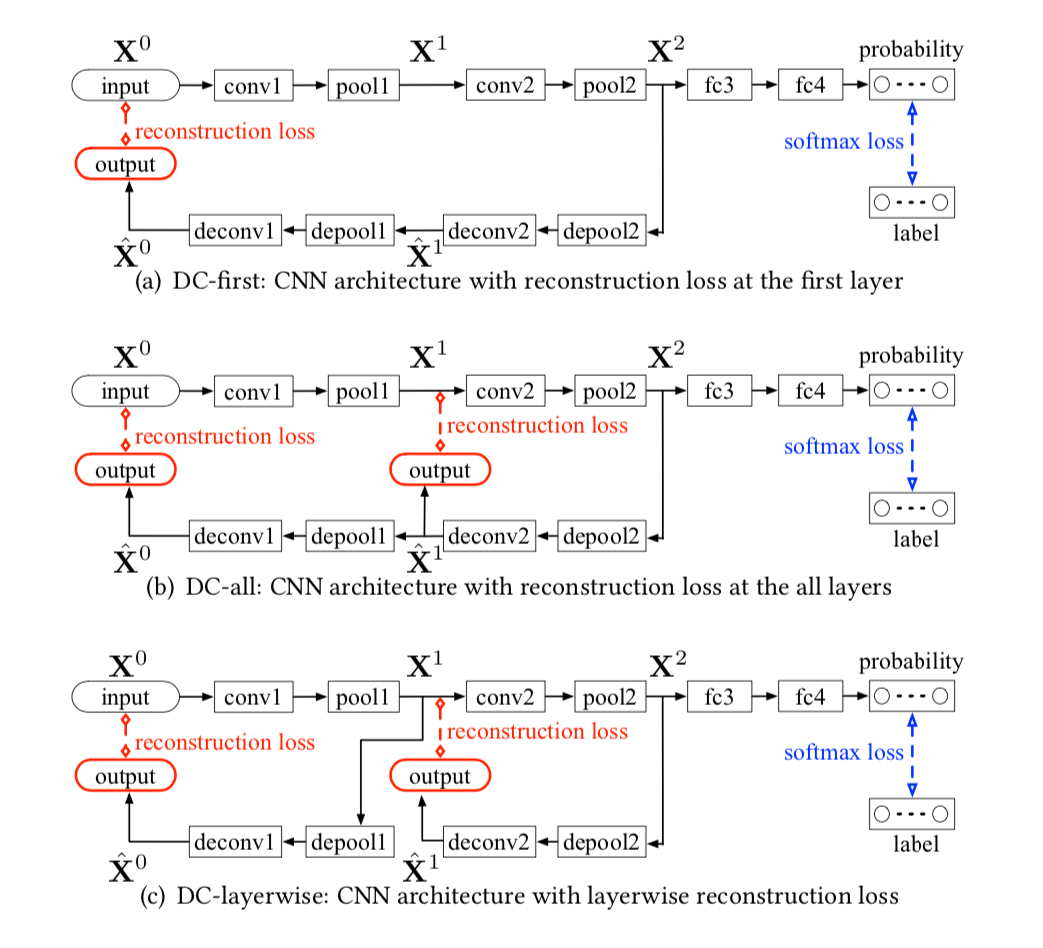}
    \caption{{\sdbn} architecture with three types of unsupervised learning augmentation \cite{sdbn}.} 
    \label{fig:sdbn_model_2}
\end{figure} 

As illustrated in Figure~\ref{fig:sdbn_model_2}, based on the input matrix $\mb{X}$ for the graphs in $\mc{G}$ (here, the subscript of $\mb{X}$ is not indicated and it can represent any graphs in $\mc{G}$), {\sdbn} proposes to apply the convolutional neural network for the graph representation learning. To be specific, the convolutional neural network used in {\sdbn} involves two operators: \textit{conv} and \textit{pool}, which can be stacked together multiple times to form a deep architecture. 

Formally, according to Figure~\ref{fig:sdbn_model_2}, the intermediate representations of the input graphs as well as the corresponding labels in the {\sdbn} can be represented with the following equations:
\begin{equation}
\begin{cases}
\mb{X}^{(1)} &= pool \left( conv \left( \mb{X}; \mb{\Theta} \right) \right);\\
\mb{X}^{(2)} &= pool \left( conv \left( \mb{X}^{(1)}; \mb{\Theta} \right) \right);\\
\mb{x} &= reshape(\mb{X}^{(2)});\\
\hat{\mb{y}} &= FC\left( \mb{x}; \mb{\Theta} \right),
\end{cases}
\end{equation}
where $reshape(\cdot)$ flattens the matrix to a matrix and $FC(\cdot)$ denotes the fully-connected layers in the model. In the above equations, $\mb{\Theta}$ denotes the involved variables in the model, which will be optimized. 

Based on the above model, for all the graph instances $G_i \in \mc{G}$, we can represent the introduced loss terms by the model as
\begin{equation}
\ell(\mb{\Theta}) = \sum_{G_i \in \mc{G}} \ell(G_i; \mb{\Theta}) = \sum_{G_i \in \mc{G}} \ell(\mb{y}_i, \hat{\mb{y}}_i),
\end{equation}
where $\mb{y}_i$ and $\hat{\mb{y}}_i$ represent the ground-truth label vector and the inferred label vector of graph $G_i$, respectively.

Meanwhile, in addition to the above loss term, {\sdbn} also incorporates the autoencoder into the model learning process via the \textit{depool} and \textit{deconv} operations. The \textit{conv} and \textit{pool} operators mentioned above compose the encoder part, whereas the \textit{deconv} and \textit{depool} operators will form the decoder part. Formally, based on the learned intermediate representation $\mb{X}^{(2)}$ of the input graph matrix $\mb{X}$, {\sdbn} computes the recovered representations as follows:
\begin{equation}
\begin{aligned}
\hat{\mb{X}}^{(1)} &= deconv \left( depool \left( \mb{X}^{(2)} \right) ; \mb{\Theta} \right);\\
\hat{\mb{X}} &= deconv \left( depool \left(\hat{\mb{X}}^{(1)} \right) ; \mb{\Theta} \right).
\end{aligned}
\end{equation}
By minimizing the difference between $\hat{\mb{X}}^{(1)}$ and ${\mb{X}}^{(1)}$, as well as the difference between $\hat{\mb{X}}$ and ${\mb{X}}$, i.e.,
\begin{equation}
\ell(\hat{\mb{X}}^{(1)}, {\mb{X}}^{(1)}) = \left\| \hat{\mb{X}}^{(1)} - {\mb{X}}^{(1)} \right\|_2^2; \mbox{ and } \ell(\hat{\mb{X}}, {\mb{X}}) = \left\| \hat{\mb{X}} - {\mb{X}} \right\|_2^2
\end{equation} 
{\sdbn} can effectively learn the involved variables in the model. As illustrated in Figure~\ref{fig:sdbn_model_2}, the decoder step can work in different manner, which will lead to different regularization terms on the intermediate representations. The performance comparison between {\isonn} and {\sdbn} is also reported in \cite{isonn}, and the readers may refer to \cite{isonn, sdbn} for more detailed information of the models and the experimental evaluation results.


\subsection{\lfer: Deep Autoencoder based Latent Feature Extraction}

Deep Autoencoder based Latent Feature Extraction ({\lfer}) initially proposed in \cite{lfer} serves an a latent feature extraction component in the final model introduced in that paper. Based on the input community real-estate POI graphs, {\lfer} aims to learn the latent representations of the POI graphs, which will be used to infer the community vibrancy scores.

\begin{figure}
	\centering
	\includegraphics[width=0.75\textwidth]{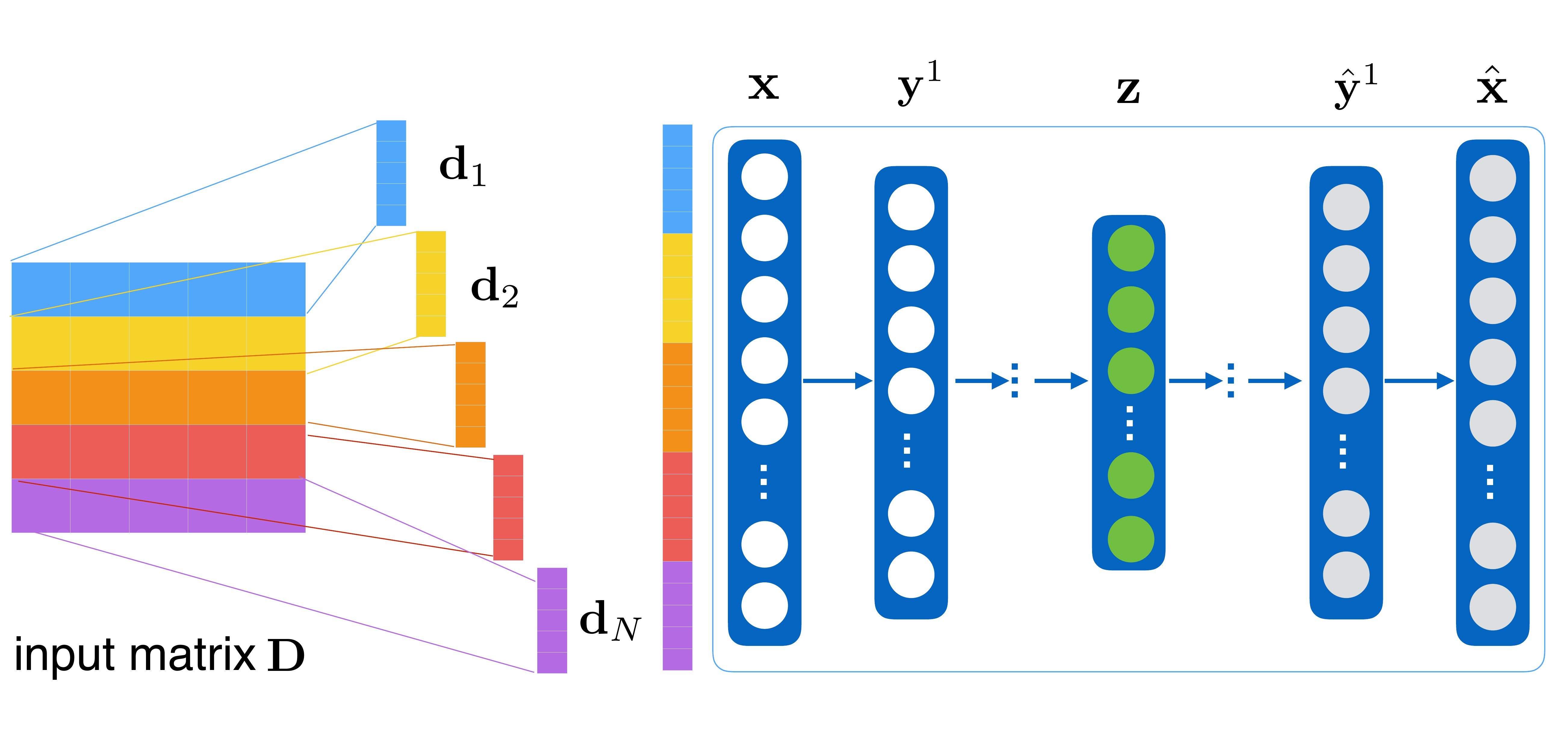}
	\caption{The {\lfer} Framework for Latent Feature Extraction \cite{lfer}.}
	\label{fig:latent_architecture}
\end{figure}

\subsubsection{Deep Autoencoder Model}

The {\lfer} model works based on the deep autoencoder actually. Autoencoder is an unsupervised neural network model, which projects the instances in original feature representations into a lower-dimensional feature space via a series of non-linear mappings. Figure~\ref{fig:latent_architecture} shows that autoencoder model involves two steps: encode and decode. The encode part projects the original feature vector to the objective feature space, while the decode step recovers the latent feature representation to a reconstruction space. In autoencoder model, we generally need to ensure that the original feature representation of instances should be as similar to the reconstructed feature representation as possible.

Formally, let $\mb{x}$ represent the original feature representation of instance $i$, and $\mb{y}^{(1)}, \mb{y}^{(2)}, \cdots, \mb{y}^{(o)}$ be the latent feature representations of the instance at hidden layers $1, 2, \cdots, o$ in the encode step respectively, the encoding result in the objective lower-dimension feature space can be represented as $\mb{z} \in \mathbb{R}^{d_z}$ with dimension $d_z$. Formally, the relationship between these vector variables can be represented with the following equations:
\vspace{-0.1cm}
\begin{equation}\begin{cases}
\mb{y}^{(1)} &= \sigma (\mb{W}^{(1)} \mb{x} + \mb{b}^{(1)}),\\
\mb{y}^{(k)} &= \sigma (\mb{W}^{(k)} \mb{y}^{{(k-1)}} + \mb{b}^{(k)}), \forall k \in \{2, 3, \cdots, o\},\\
\mb{z} &= \sigma (\mb{W}^{{(o+1)}} \mb{y}^{(o)} + \mb{b}^{(o+1)}).
\end{cases}\end{equation}

Meanwhile, in the decode step, the input will be the latent feature vector $\mb{z}$ (i.e., the output of the encode step), and the final output will be the reconstructed vector $\hat{\mb{x}}$. The latent feature vectors at each hidden layers can be represented as $\hat{\mb{y}}^{(o)}, \hat{\mb{y}}^{(o-1)}, \cdots, \hat{\mb{y}}^{(1)}$. The relationship between these vector variables can be denoted as
\begin{equation}\begin{cases}
\hat{\mb{y}}^{(o)} &= \sigma (\hat{\mb{W}}^{(o+1)} \mb{z} + \hat{\mb{b}}^{(o+1)}),\\
\hat{\mb{y}}^{(k-1)} &= \sigma (\hat{\mb{W}}^{(k)} \hat{\mb{y}}^{(k)} + \hat{\mb{b}}^{(k)}), \forall k \in \{2, 3, \cdots, o\},\\
\hat{\mb{x}} &= \sigma(\hat{\mb{W}}^{(1)} \hat{\mb{y}}^{(1)} + \hat{\mb{b}}^{(1)}).
\end{cases}\end{equation}

In the above equations, $\mb{W}$ and $\mb{b}$ with different subscripts denote the weight matrices and bias terms to be learned in the model. The objective of the autoencoder model is to minimize the loss between the original feature vector $\mb{x}$ and the reconstructed feature vector $\hat{\mb{x}}$. Formally, the loss term can be represented as
\begin{equation}
\ell(\bs{\Theta}) = \ell(\mb{x}, \hat{\mb{x}}; \bs{\Theta}) = \left \| {\mb{x}} - \hat{\mb{x}}\right\|_2^2,
\end{equation}
where $\bs{\Theta}$ denotes the variables involved in the autoencoder model.

\subsubsection{Latent Representation Learning}

{\lfer} proposes to learn the community allocation information for the vibrancy inference and ranking. Formally, \textit{spatial structure} denotes the distribution of POIs inside the community, e.g., a grocery store lies between two residential buildings; a school is next to the police office. The \textit{Spatial structure} can hardly be represented with explicit features extracted before, and {\lfer} proposes to represent them with a set of latent feature vectors extracted from the \textit{geographic distance graph} and the \textit{mobility connectivity graph} defined in the previous subsection. The autoencoder model is applied here for the latent feature extraction. 

Autoencoder model has been applied to embed the graph data into lower-dimensional spaces in many of the research works, which will obtain a latent feature representation for the nodes inside the graph. Different from these works, instead of calculating the latent feature for the POI categories inside the communities, {\lfer} aims at obtaining the latent feature vector for the whole community, i.e., embedding the graph as one latent feature vector. 

As shown in Figure~\ref{fig:latent_architecture}, {\lfer} transforms the matrix of the \textit{graphical distance graph} (involving the POI categories) $\mb{D}$ into a vector, which can be denoted as
\begin{equation}
\mb{d} = reshape(\mb{D}) \in \mathbbm{R}^{|\mc{V}|^2 \times 1}.
\end{equation}
Vector $\mb{d}$ will be used as the input feeding into the autoencoder model. The latent embedding feature vector of $\mb{d}$ can be represented as $\mb{z}_D$ (i.e., the vector $\mb{z}$ as introduced in the autoencoder model in the previous section), which depicts the layout information of POI categories in the community in terms of the geographical distance. Besides the static layout based on \textit{geographic distance graph}, the spatial structure of the POIs in the communities can also be revealed indirectly through the human mobility. For a pair of POI categories which are far away geographically, if people like to go between them frequently, it can display another type of structure of the POIs in terms of their functional correlations. Via the multiple fully connected layers, {\lfer} will project such learned features to the objective vibrancy scores of the community. We will not introduce the model learning part here, since it also involves the ranking models and explicit feature engineering works, which is not closely related to the topic of this paper. The readers may refer to \cite{lfer} for detailed description about the model and its learning process. In addition, autoencoder (i.e., the base model of {\lfer}) is also compared against {\isonn}, whose results are reported in \cite{isonn}.




\section{Graph Neural Networks for Giant Networks}\label{sec:giant_network}

In this section, we will introduce the graph neural networks proposed for the representation learning tasks on giant networks instead. Formally, we can represent the giant network instance to be studied in this section as $G = (\mc{V}, \mc{E})$, where $\mc{V}$ and $\mc{E}$ denote the sets of nodes and links in the network, respectively. Different from the small graph data studied in Section~\ref{sec:small_graph}, the nodes in the giant network $G$ are partially annotated with labels instead. Formally, we can represent the set of labeled nodes as $\mc{V}_{\textsc{l}} = \{(v_{\textsc{l}, 1}, \mb{y}_{\textsc{l}, 1}), (v_{\textsc{l}, 2}, \mb{y}_{\textsc{l}, 2}), \cdots, (v_{\textsc{l}, m}, \mb{y}_{\textsc{l}, m})\}$, where $v_{\textsc{l}, i} \in \mc{V}, \forall i \in \{1, 2, \cdots, m\}$ and $\mb{y}_{\textsc{l}, i}$ denotes its label vector; whereas the remaining unlabeled nodes can be represented as $\mc{V}_{\textsc{u}} = \mc{V} \setminus \mc{V}_{\textsc{l}}$. In the case where all the involved network nodes are labeled, we will have $\mc{V}_{\textsc{l}} = \mc{V}$ and $\mc{V}_{\textsc{u}} = \emptyset$, which will be a special learning scenario of the general partial-labeled learning setting as studied in this paper. The giant network oriented graph neural networks aim at learning a mapping, i.e., $f: \mc{V} \to \mathbbm{R}^{d_h}$, to obtain the feature vector representations of the nodes in the network, which can be utilized to infer their labels. To be more specific, the models to be introduced in this section include {\gcn} \cite{gcn}, {\gat} \cite{gat}, {\dif} \cite{dif}, {\gnl} \cite{gnl}, {\sage} \cite{sage} and {\segen} \cite{segen}.


\subsection{{\gcn}: Graph Convolutional Network}\label{subsec:gcn}

Graph convolutional network ({\gcn}) initially proposed in \cite{gcn} introduces a \textit{spectral graph convolution} operator for the graph data representation learning, which also provides several different approximations of the operator to encode both the graph structure and features of the nodes. {\gcn} works well for the partially labeled giant networks, and the learned node representations can be effectively applied for the node classification task.

\subsubsection{Spectral Graph Convolution}

Formally, given an input network $G = (\mc{V}, \mc{E})$, its network structure information can be denoted as an adjacency matrix $\mb{A} = \{0, 1\}^{|\mc{V}| \times |\mc{V}|}$. The corresponding normalized graph Laplacian matrix can be denoted as $\mb{L} = \mb{I} - \mb{D}^{-\frac{1}{2}} \mb{A} \mb{D}^{-\frac{1}{2}} = \mb{U} \mb{\Lambda} \mb{U}^\top$, where $\mb{D}$ is a diagonal matrix with entries $\mb{D}(i,i) = \sum_j \mb{A}(i,j)$ on its diagonal and $\mb{I} = diag(\mb{1}) \in \{0, 1\}^{|\mc{V}| \times |\mc{V}|}$ is an identity matrix with ones on its diagonal. The eigen-decomposition of matrix $\mb{L}$ can be denoted as $\mb{L} = \mb{U} \mb{\Lambda} \mb{U}^\top$, where $\mb{U}$ denotes the eigen-vector matrix and diagonal matrix $\mb{\Lambda}$ has eigen-values on its diagonal.

The \textit{spectral convolution} operator defined on network $G$ in {\gcn} is denoted as a multiplication of an input signal vector $\mb{x} \in \mathbbm{R}^{d_x}$ with a filter $\mb{g}_{\bs{\theta}} = diag(\bs{\theta})$ (parameterized by variable vector $\bs{\theta} \in \mathbbm{R}^{d_{\theta}})$ in the Fourier domain as follows:
\begin{equation}\label{equ:spectral_convolution}
\mb{g}_{\bs{\theta}} * \mb{x} = \mb{U} \mb{g}_{\bs{\theta}} \mb{U}^{\top} \mb{x},
\end{equation}
where notation $\mb{U}^{\top} \mb{x}$ is defined as the graph Fourier transform of $\mb{x}$ and $\mb{g}_{\bs{\theta}}$ can be understood as a function on the eigen-values, i.e., $\mb{g}_{\bs{\theta}}(\mb{\Lambda})$.

According to Equ.~(\ref{equ:spectral_convolution}), the computation cost of the term on the right-hand-side will be $\mc{O}(|\mc{V}|^2)$. For the giant networks involving millions even billions of nodes, the computation of the graph convolution term will become infeasible, not to mention the eigen-decomposition of the Laplacian matrix $\mb{L}$ defined before. Therefore, to resolve such a problem, \cite{gcn} introduces an approximation of the filter function $\mb{g}_{\bs{\theta}}(\mb{\Lambda})$ by a truncated expansion in terms of the Chebyshev polynomial $T_k(\cdot)$ up to the $K_{th}$ order as follows:
\begin{equation}\label{equ:spectral_convolution_approx}
\mb{g}_{\bs{\theta}} * \mb{x} \approx \sum_{k = 0}^K \bs{\theta}(k) T_k(\tilde{\mb{L}}) \mb{x},
\end{equation}
where $\tilde{\mb{L}} = \frac{2}{\lambda_{max}} \mb{L} - \mb{I}$ and $\lambda_{max}$ is the largest eigen-value in matrix $\mb{\Lambda}$. Vector $\bs{\theta} \in \mathbbm{R}^k$ is a vector of Chebyshev coefficients. Noticing that the computational complexity of the term on the right-hand-side is $\mc{O}(|\mc{E}|)$, i.e., linear in terms of the edge numbers, which will be lower than that of Equ.~(\ref{equ:spectral_convolution}) introduced before.

\begin{figure}
	\centering
	\includegraphics[width=0.75\textwidth]{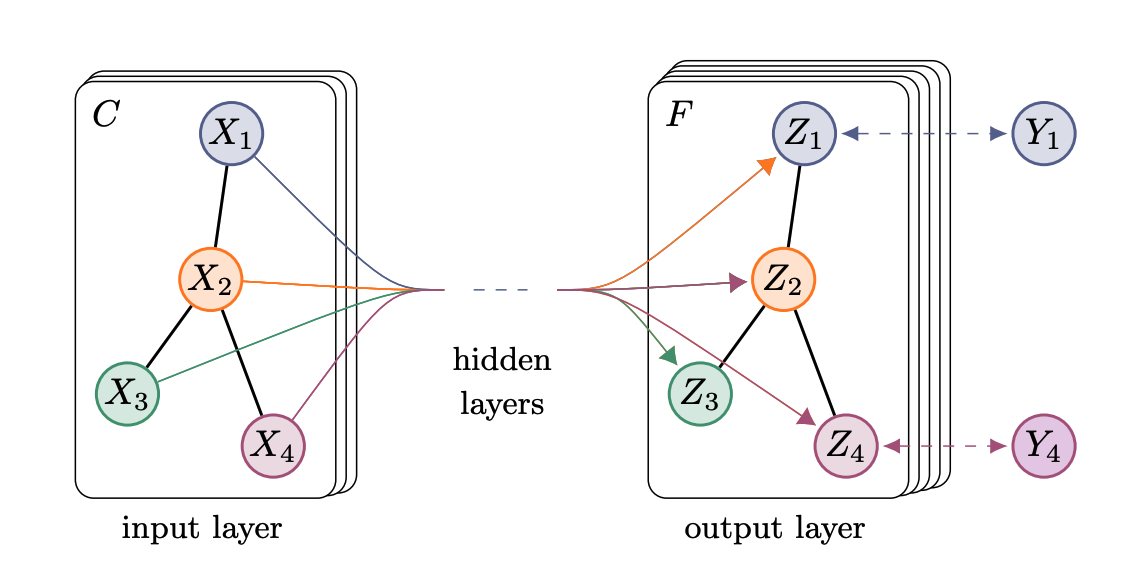}
	\caption{Schematic depiction of multi-layer {\gcn} for semisupervised learning with $C$ input channels and $F$ feature maps in the output layer \cite{gcn}.}
	\label{fig:gcn_architecture}
\end{figure}

\subsubsection{Graph Convolution Approximation}

As proposed in \cite{gcn}, Equ.~(\ref{equ:spectral_convolution_approx}) can be further simplified by setting $K=1$, $\lambda_{max} = 2$, $\bs{\theta}(0) = - \bs{\theta}(1) = \theta$, which will reduce the right hand-term of Equ.~(\ref{equ:spectral_convolution_approx}) approximately as follows:
\begin{equation}
\begin{aligned}
\mb{g}_{\bs{\theta}} * \mb{x}  \approx \bs{\theta}(0) \mb{x} +   \bs{\theta}(1) \tilde{\mb{L}} \mb{x} &= {\theta} (\mb{I} + \mb{D}^{-\frac{1}{2}} \mb{A} \mb{D}^{-\frac{1}{2}}) \mb{x}\\
&=  {\theta} (\tilde{\mb{D}}^{-\frac{1}{2}} \tilde{\mb{A}} \tilde{\mb{D}}^{-\frac{1}{2}}) \mb{x},
\end{aligned}
\end{equation}
where $\tilde{\mb{A}} = \mb{A} + \mb{I}$ and $\tilde{\mb{D}}$ is the diagonal matrix defined on $\tilde{\mb{A}}$ instead.

As illustrated in Figure~\ref{fig:gcn_architecture}, in the case when there exist $C$ input channels, i.e., the input will be a matrix $\mb{X} \in \mathbbm{R}^{|\mc{V}| \times C}$, and $F$ different filters defined above, the learned graph convolution feature representations will be 
\begin{equation}\label{equ:gcn_update}
\begin{aligned}
\mb{Z} &= (\tilde{\mb{D}}^{-\frac{1}{2}} \tilde{\mb{A}} \tilde{\mb{D}}^{-\frac{1}{2}})  \mb{X} \mb{W},\\
&= \hat{\mb{A}}  \mb{X} \mb{W}.
\end{aligned}
\end{equation}
where matrix $\hat{\mb{A}} = \tilde{\mb{D}}^{-\frac{1}{2}} \tilde{\mb{A}} \tilde{\mb{D}}^{-\frac{1}{2}}$ can be pre-computed in advance. Matrix $\mb{W} \in \mathbbm{R}^{C \times F}$ is the filter parameter matrix and $\mb{Z} \in \mathbbm{R}^{|\mc{V}| \times F}$ will be the learned convolved representations of all the nodes. The computational time complexity of the operation will be $\mc{O}(|\mc{E}| F C)$.

\subsubsection{Deep Graph Convolutional Network Learning}\label{subsec:gcn_training}

The {\gcn} model can have a deeper architecture by involving multiple graph convolution operators defined in the previous sections. For instance, the {\gcn} model with two layers can be represented with the following equations:
\begin{equation}\label{equ:deep_gcn}
\begin{cases}
&\mbox{Layer 1: } \mb{Z} = \mbox{ReLU}\left( \hat{\mb{A}} \mb{X} \mb{W}_1 \right);\\
&\mbox{Output Layer: } \hat{\mb{Y}} = \mbox{softmax}\left( \hat{\mb{A}} \mb{Z} \mb{W}_2 \right).
\end{cases}
\Rightarrow
\hat{\mb{Y}} = \mbox{softmax}\left( \hat{\mb{A}} \mbox{ReLU}\left( \hat{\mb{A}} \mb{X} \mb{W}_1 \right) \mb{W}_2 \right).
\end{equation}
In the above equation, matrices $\mb{W}_1$ and $\mb{W}_2$ are the involved variables in the model. ReLU is used as the activation function for the hidden layer 1, and softmax function is used for the output result normalization. By comparing the inferred labels, i.e., $\hat{\mb{Y}}$, of the labeled instances against their ground-truth labels, i.e., ${\mb{Y}}$, the model variables can be effectively learned by minimizing the following loss function:
\begin{equation}
\ell(\bs{\Theta}) = - \sum_{v_i \in \mc{V}_{\textsc{L}}} \sum_{j=1}^{d_y} \mb{Y}(i, j) \log \hat{\mb{Y}}(i,j),
\end{equation}
where $\bs{\Theta}$ covers all the variables in the model.

For representation simplicity, node subscript is used as its corresponding index in the label matrix $\mb{Y}$. Notation $d_y$ denotes the number of labels in the studied problem, and $d_y = 2$ for the traditional binary classification tasks. The readers can also refer to \cite{gcn} for detailed information of the {\gcn} model.


\subsection{{\gat}: Graph Attention Network}

\begin{figure}
	\centering
	\includegraphics[width=0.9\textwidth]{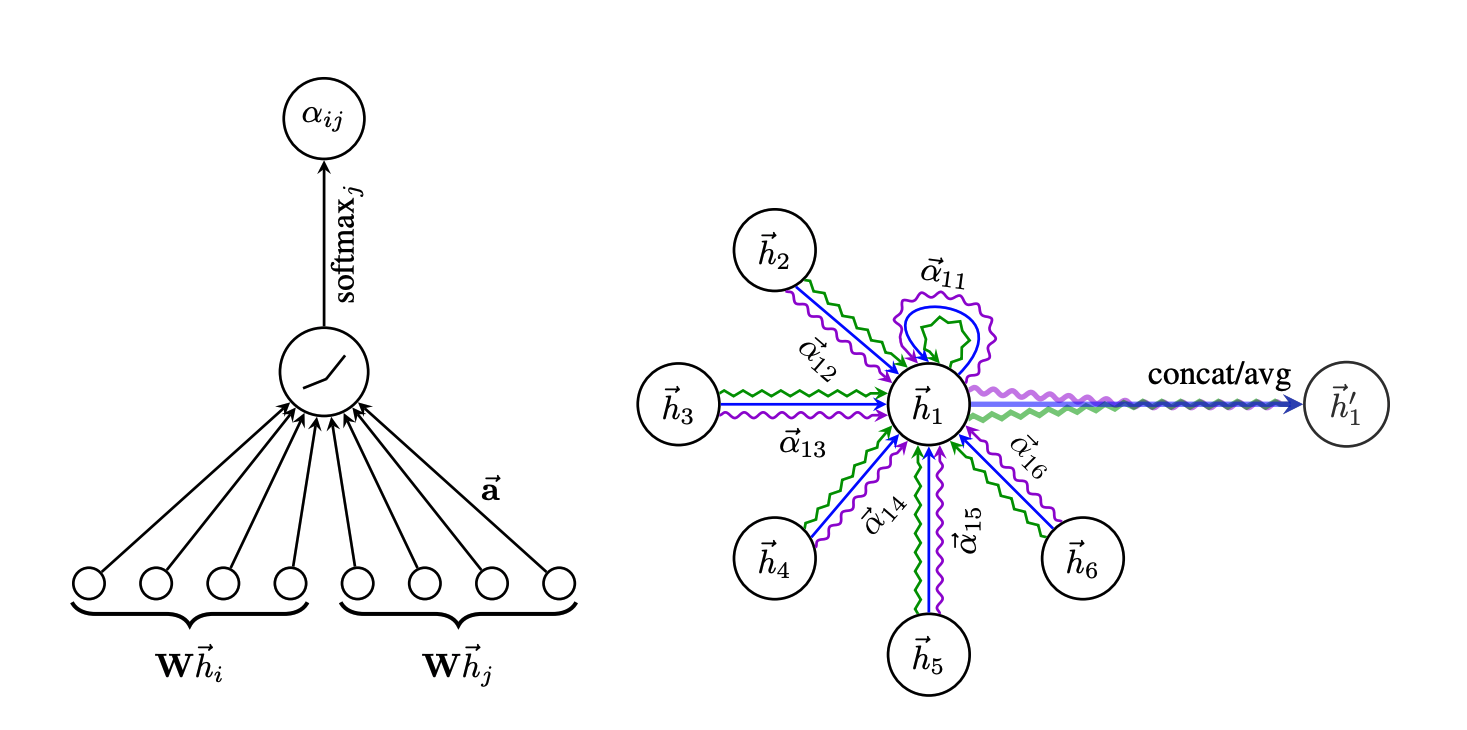}
	\caption{Schematic depiction of {\gat} with multi-head attention for the node representation update \cite{gat}.}
	\label{fig:gat_architecture}
\end{figure}

Graph attention network ({\gat}) initially proposed in \cite{gat} can be viewed as an extension of {\gcn}. In updating the nodes' representations, instead of assigning the neighbors with fixed weights, i.e., values in matrix $\hat{\mb{A}}$ in Equ.~(\ref{equ:gcn_update}) and Equ.~(\ref{equ:deep_gcn}), {\gat} introduces an attention mechanism to compute the weights based on the representations of the target node as well as its neighbors.

\subsubsection{Graph Attention Coefficient Computation}

Formally, given an input network $G = (\mc{V}, \mc{E})$ and the raw features of the nodes, the node features can be denoted as a matrix $\mb{X} \in \mathbbm{R}^{|\mc{V}| \times d_x}$, where $d_x$ denotes the dimension of the node feature vectors. Furthermore, for node $v_i \in \mc{V}$, its feature vector can also be represented as $\mb{x}_i = \mb{X}(i,:)$ for simplicity. Without considerations about the network structures, via a mapping matrix $\mb{W} \in \mathbbm{R}^{d_x \times d_h}$, the nodes can be projected to their representations in the hidden layer. Meanwhile, to further incorporate the network structure information into the model, based on the network structure, the neighbor set of node $v_i$ can be denoted as $\Gamma(v_i) = \{v_j | v_j \in \mc{V} \land (v_i, v_j) \in \mc{E}\} \cup \{v_i\}$, where $\{v_i\}$ is also added and treated as the \textit{self-neighbor}. As illustrated in Figure~\ref{fig:gat_architecture}, {\gat} proposes to compute the attention coefficient between nodes $v_i$ and $v_j$ (if $v_j \in \Gamma(v_i$)) as follows:
\begin{equation}
e_{i,j} = \mbox{LeakyReLU} \left( \mb{a}^\top \left( \mb{W} \mb{x}_i \sqcup \mb{W} \mb{x}_j \right) \right),
\end{equation}
where $\mb{a} \in \mathbbm{R}^{2d_h}$ is a variable vector for weighted sum of the entries in vector $\mb{W} \mb{x}_i \sqcup \mb{W} \mb{x}_j$ and $\sqcup$ denotes the concatenation operator of two vectors. LeakyReLU function is added here mainly for the model learning considerations.

To further normalize the coefficients among all the neighbors, {\gat} further adopts the softmax function based on the coefficients defined above. Formally, the final computed weight between nodes $v_i$ and $v_j$ can be denoted as
\begin{equation}
\begin{aligned}
\alpha_{i,j} &= softmax (e_{i,j}) \\
&= \frac{\exp(e_{i,j})}{\sum_{v_k \in \Gamma(v_i)} \exp(e_{i,k})}\\
&= \frac{\exp\left( \mbox{LeakyReLU} \left( \mb{a}^\top \left( \mb{W} \mb{x}_i \sqcup \mb{W} \mb{x}_j \right) \right) \right)}{\sum_{v_k \in \Gamma(v_i)} \exp\left( \mbox{LeakyReLU} \left( \mb{a}^\top \left( \mb{W} \mb{x}_i \sqcup \mb{W} \mb{x}_k \right) \right) \right)}.
\end{aligned}
\end{equation}

\subsubsection{Representation Update via Neighborhood Aggregation}

{\gat} effectively update the nodes' representations by aggregating the information from their neighbors (including the \textit{self-neighbor}). Formally, the learned hidden representation of node $v_i$ can be represented as
\begin{equation}\label{equ:representation_update}
\mb{h}_i = \sigma \left( \sum_{v_j \in \Gamma(v_i)} \alpha_{i,j} \mb{W} \mb{x}_j \right).
\end{equation}

{\gat} can be in a deeper architecture by involving multiple attentive node representation updating. In the deep architecture, for the upper layers, the representation vector $\mb{h}_i$ will be treated as the inputs feature vector instead, and we will not over-elaborate that here.

\subsubsection{Multi-Head Attention Aggregation}

As introduced in \cite{gat}, to stabilize the learning process of the model, {\gat} can be further extended to include the multi-head attention as illustrated in Figure~\ref{fig:gat_architecture}. Specifically, let $K$ denote the number of involved attention mechanisms. Based on each attention mechanism, the learned representations of node $v_i$ based on the above descriptions (i.e., Equ.~(\ref{equ:representation_update})) can be denoted as $\mb{h}_i^{(1)}$, $\mb{h}_i^{(2)}$, $\cdots$, $\mb{h}_i^{(K)}$, respectively. By further aggregating such learned representations together, the ultimate learned representation of node $v_i$ can be denoted as
\begin{equation}
\mb{h}_i' = \mbox{Aggregate}(\mb{h}_i^{(1)}, \mb{h}_i^{(2)}, \cdots, \mb{h}_i^{(K)}).
\end{equation}

Several different aggregation function is tried in \cite{gat}, including \textit{concatenation} and \textit{average}:
\begin{itemize}
\item \textit{Concatenation}: 
\begin{equation}
\mb{h}_i' = \bigsqcup_{k=1}^K \sigma \left(  \sum_{v_j \in \Gamma(v_i)} \alpha^{(k)}_{i,j} \mb{W}^{(k)} \mb{x}_j \right).
\end{equation}

\item \textit{Average}: 
\begin{equation}
\mb{h}_i' = \sigma\left( \frac{1}{K} \sum_{k=1}^K \sum_{v_j \in \Gamma(v_i)} \alpha^{(k)}_{i,j} \mb{W}^{(k)} \mb{x}_j \right).
\end{equation}
\end{itemize}
The learning process of the {\gat} model is very similar to that of {\gcn} introduced in Section~\ref{subsec:gcn_training}, and we will not introduce that part again here. The readers can also refer to \cite{gat} for more detailed descriptions about the model as well as its experimental performance.


\subsection{{\dif}: Deep Diffusive Neural Network}

Deep diffusive network ({\dif}) model initially introduced in \cite{dif} aims at modeling the diverse connections in heterogeneous information networks, which contains multiple types of nodes and links. {\dif} is based on a new type of neuron, namely \textit{gated diffusive unit} ({\gdu}), which can be extended to incorporate the inputs from various groups of neighbors. 


\subsubsection{Model Architecture}

\begin{figure}[t]
\minipage{0.48\textwidth}
  \includegraphics[width=\linewidth]{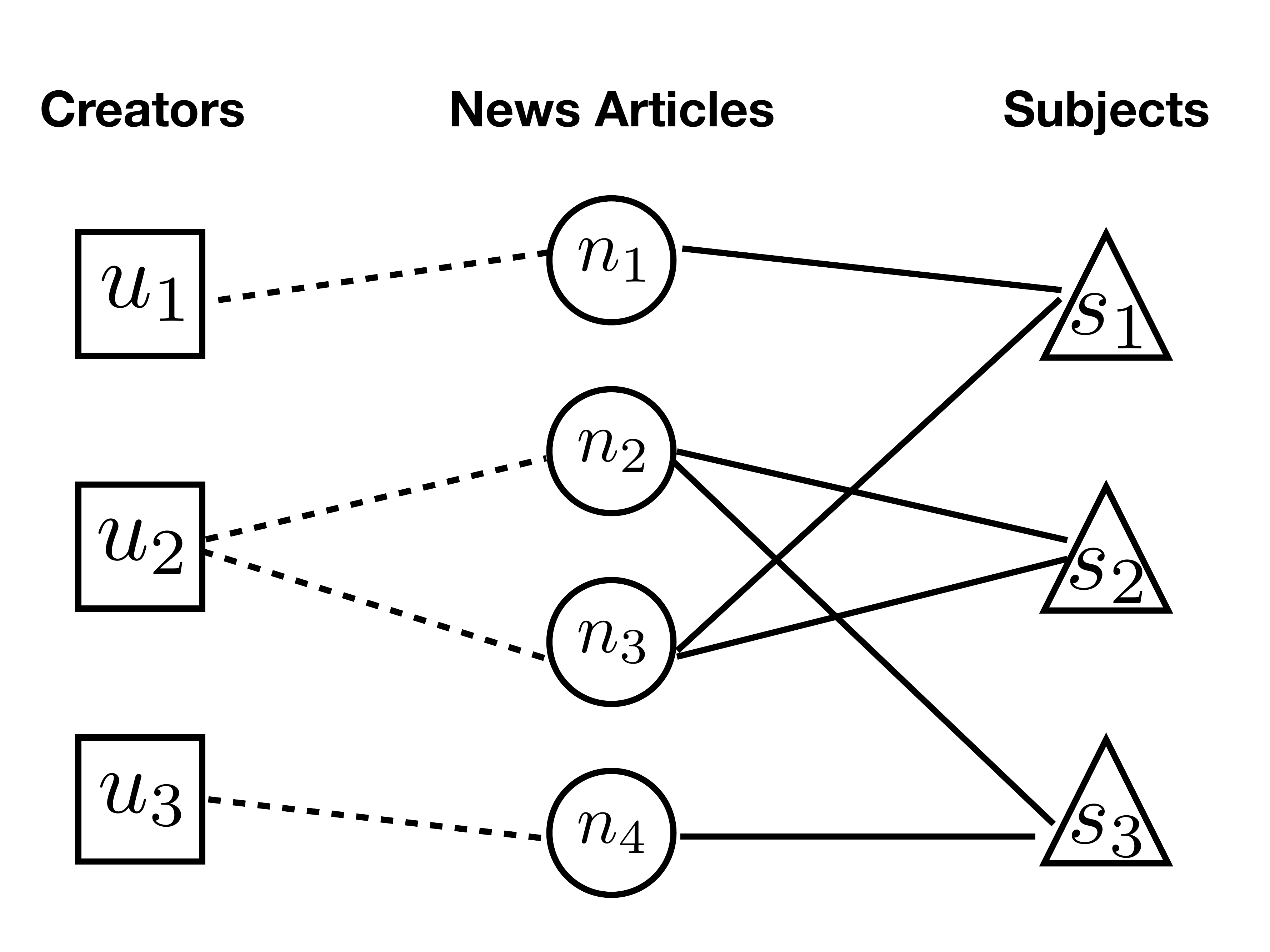}
  \caption{An example of the news augmented heterogeneous social network}\label{fig:heterogeneous_network_example}
\endminipage\hfill
\minipage{0.5\textwidth}
  \includegraphics[width=\linewidth]{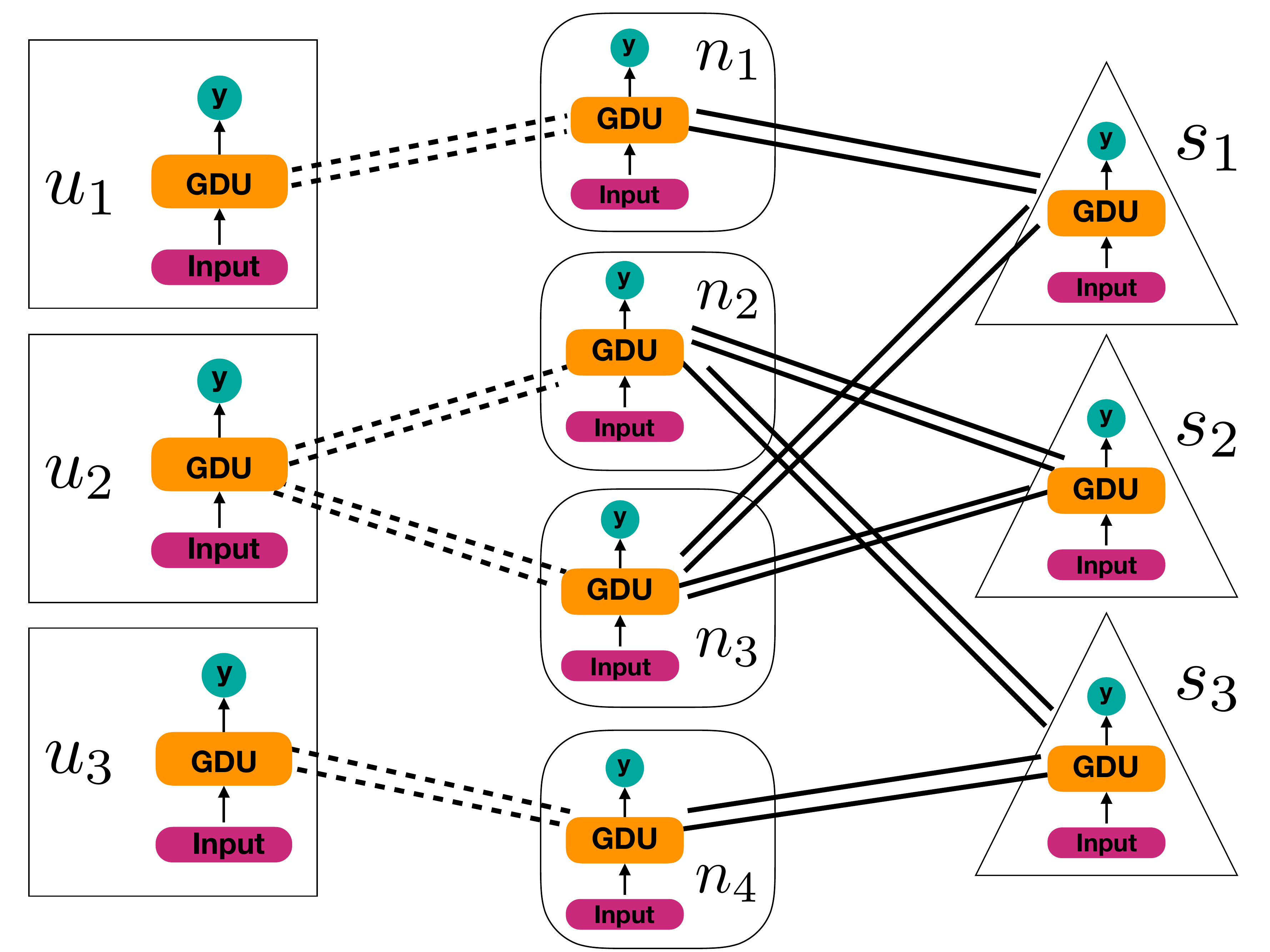}
  \caption{The {\dif} model architecture build for the input heterogeneous social network}\label{fig:dif_architecture}
\endminipage
\end{figure}

Given a heterogeneous input network $G = (\mc{V}, \mc{E})$, the node set $\mc{V}$ in the network can be divided into multiple subsets depending on their node types. It is similar for the links in set $\mc{E}$. Here, for the representation simplicity, we will follow the news augmented heterogeneous social network example illustrated in \cite{dif} when introducing the model. As illustrated in Figure~\ref{fig:heterogeneous_network_example}, there exist three different types of nodes (i.e., \textit{creator}, \textit{news article} and \textit{subject}) and two different types of links (i.e., the \textit{creator-article link} and \textit{article-subject link}) in the network. Formally, the node set can be categories into three subsets, i.e., $\mc{V} = \mc{U} \cup \mc{N} \cup \mc{S}$, and the link set can be categorized into two subsets, i.e., $\mc{E} = \mc{E}_{u,n} \cup \mc{E}_{n,s}$.

For each node in the network, e.g., $v_i \in \mc{V}$, its extracted raw feature vector can be denoted as $\mb{x}_i$. As introduced at the beginning of Section~\ref{sec:giant_network}, in many cases, the network is partially labeled. Formally, the label vector of node $v_i$ is represented as $\mb{y}_i$. For each nodes in the input network, {\dif} utilizes one {\gdu} (which will be introduced in the following subsection) to model its representations and the connections with other neighboring nodes. For instance, based on the input network in Figure~\ref{fig:heterogeneous_network_example}, its corresponding {\dif} model architecture can be represented in Figure~\ref{fig:dif_architecture}. Via the {gdu} neuron unit, {\dif} can effectively project the node inputs to their corresponding labels. The parameters involved in the {\dif} model can be effectively trained based on the labeled nodes via the back propagation algorithm. In the following two subsections, we will introduce the detailed information about {\gdu} as well as the {\dif} model training.



\subsubsection{Gated Diffusive Unit}\label{subsec:gdu}

\begin{figure}[t]
\minipage{\textwidth}
\centering
  \includegraphics[width=0.5\linewidth]{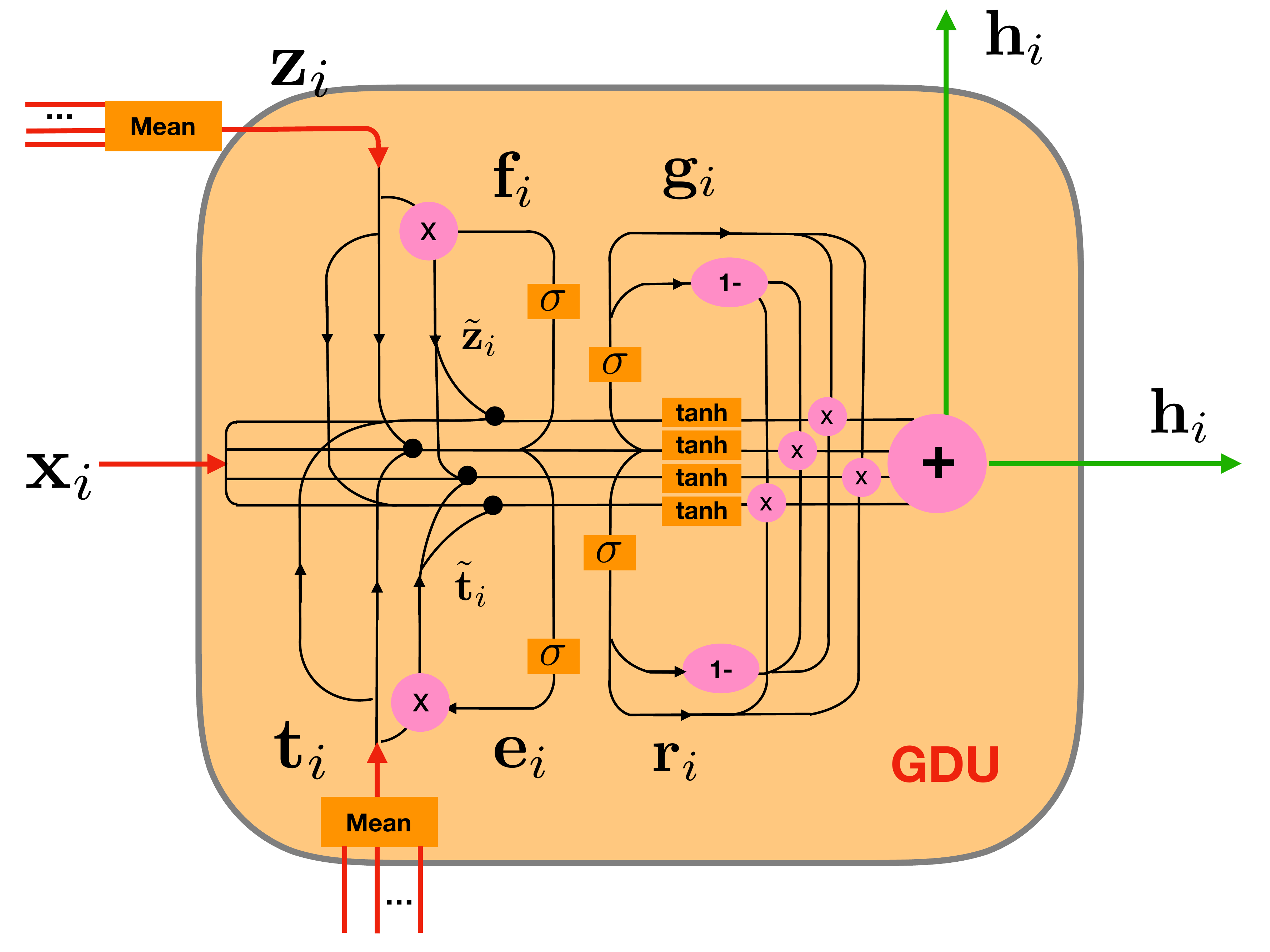}
  \caption{An illustration of the gated diffusive unit ({\gdu}).}\label{fig:dif_gdu}
\endminipage
\end{figure}

To introduce the {\gdu} neuron, we can take news article nodes as an example here. Formally, among all the inputs of the {\gdu} model, $\mb{x}_i$ denotes the extracted feature vector for news articles, $\mb{z}_i$ represents the input from other {\gdu}s corresponding to subjects, and $\mb{t}_i$ represents the input from other {\gdu}s about creators. Considering that the {\gdu} for each news article may be connected to multiple {\gdu}s of subjects and creators, the $mean(\cdot)$ of the outputs from the {\gdu}s corresponding to these subjects and creators will be computed as the inputs $\mb{z}_i$ and $\mb{t}_i$ instead respectively, which is also indicated by the {\gdu} architecture illustrated in Figure~\ref{fig:dif_gdu}. For the inputs from the subjects, {\gdu} has a gate called the ``forget gate'', which may update some content of $\mb{z}_i$ to forget. The forget gate is important, since in the real world, different news articles may focus on different aspects about the subjects and ``forgetting'' part of the input from the subjects is necessary in modeling. Formally, the ``forget gate'' together with the updated input can be represented as
\begin{equation}\label{equ:gdu_gate1}
\tilde{\mb{z}}{i} = \mb{f}_i \otimes \mb{z}_i, \mbox{ where  }\mb{f}_i = \sigma \left( \mb{W}_f \left[\mb{x}_i^\top, \mb{z}_i^\top, \mb{t}_i^\top \right]^\top \right).
\end{equation}
Here, operator $\otimes$ denotes the entry-wise product of vectors and $\mb{W}_f$ represents the variable of the forget gate in {\gdu}.

Meanwhile, for the input from the creator nodes, a new node-type ``adjust gate'' is introduced in {\gdu}. Here, the term ``adjust'' models the necessary changes of information between different node categories (e.g., from creators to articles). Formally, the ``adjust gate'' as well as the updated input can be denoted as
\begin{align}\label{equ:gdu_gate2}
\tilde{\mb{t}}_i = \mb{e}_i \otimes \mb{t}_i, \mbox{ where  }\mb{e}_i = \sigma \left( \mb{W}_e \left[\mb{x}_i^\top, \mb{z}_i^\top, \mb{t}_i^\top \right]^\top \right),
\end{align}
where $\mb{W}_e$ denotes the variable matrix in the adjust gate.

{\gdu} allows different combinations of these input/state vectors, which are controlled by the selection gates $\mb{g}_i$ and $\mb{r}_i$ respectively. Formally, the final output of {\gdu} will be
\begin{equation}\label{equ:gdu_update}
\begin{aligned}
\mb{h}_i &= \mb{g}_i \otimes \mb{r}_i \otimes \tanh \left(\mb{W}_u [\mb{x}_i^\top, \tilde{\mb{z}}_{i}^\top, \tilde{\mb{t}}_i^\top]^\top \right)\\
&+ (\mb{1} - \mb{g}_i) \otimes \mb{r}_i \otimes \tanh \left(\mb{W}_u [\mb{x}_i^\top, {\mb{z}}_i^\top, \tilde{\mb{t}}_i^\top]^\top \right)\\
&+ \mb{g}_i \otimes (\mb{1} - \mb{r}_i) \otimes \tanh \left(\mb{W}_u [\mb{x}_i^\top, \tilde{\mb{z}}_i^\top, {\mb{t}}_i^\top]^\top \right)\\
&+ (\mb{1} - \mb{g}_i) \otimes (\mb{1} - \mb{r}_i) \otimes \tanh \left(\mb{W}_u [\mb{x}_i^\top, {\mb{z}}_i^\top, {\mb{t}}_i^\top]^\top \right),
\end{aligned}
\end{equation}
where $\mb{g}_i = \sigma ( \mb{W}_g \left[\mb{x}_i^\top, \mb{z}_i^\top, \mb{t}_i^\top \right]^\top )$, and $\mb{r}_i = \sigma ( \mb{W}_r \left[\mb{x}_i^\top, \mb{z}_i^\top, \mb{t}_i^\top \right]^\top )$, and term $\mb{1}$ denotes a vector filled with value $1$. Operators $\oplus$ and $\ominus$ denote the entry-wise addition and minus operation of vectors. Matrices $\mb{W}_u$, $\mb{W}_g$, $\mb{W}_r$ represent the variables involved in the components. Vector $\mb{h}_i$ will be the output of the {\gdu} model.

The introduced {\gdu} model also works for both the news subjects and creator nodes in the network. When applying the {\gdu} to model the states of the subject/creator nodes with two input only, the remaining input port can be assigned with a default value (usually vector $\mb{0}$). In the following section, we will introduce how to learn the parameters involved in the {\dif} model for concurrent inference of multiple nodes.


\subsubsection{{\dif} Model Learning}

In the {\dif} model as shown in Figure~\ref{fig:dif_architecture}, based on the output state vectors of news articles, news creators and news subjects, the framework will project the feature vectors to their labels. Formally, given the state vectors $\mb{h}_{n,i}$ of news article $n_i$, $\mb{h}_{u,j}$ of news creator $u_j$, and $\mb{h}_{s,l}$ of news subject $s_l$, their inferred labels can be denoted as vectors $\mb{y}_{n,i}, \mb{y}_{u,j}, \mb{y}_{s,l} \in \mathcal{R}^{|\mathcal{Y}|}$ respectively, which can be represented as
\begin{equation}
\begin{cases}
\mb{y}_{n,i} &= softmax\left( \mb{W}_n \mb{h}_{n,i} \right),\\
\mb{y}_{u,j} &= softmax\left( \mb{W}_u \mb{h}_{u,j} \right),\\
\mb{y}_{s,l} &= softmax\left( \mb{W}_s \mb{h}_{s,l} \right).
\end{cases}
\end{equation}
where $\mb{W}_u$, $\mb{W}_n$ and $\mb{W}_s$ define the weight variables projecting state vectors to the output vectors.

Meanwhile, based on the news articles in the training set $\mathcal{T}_n \subset \mathcal{N}$ with the ground-truth label vectors $\{\hat{\mb{y}}_{n,i}\}_{n_i \in \mathcal{T}_n}$, the loss function of the framework for news article label learning are defined as the cross-entropy between the prediction results and the ground truth:
\begin{align}
\ell(\mathcal{T}_n; \mb{\Theta}) &= - \sum_{n_i \in \mathcal{T}_n}  \sum_{k=1}^{|\mathcal{Y}|}\hat{\mb{y}}_{n,i}(k) \log {\mb{y}}_{n,i}(k).
\end{align}
Similarly, the loss terms introduced by news creators and subjects based on training sets $\mathcal{T}_u \subset \mathcal{U}$ and $\mathcal{T}_s \subset \mathcal{S}$ can be denoted as
\begin{align}
\ell(\mathcal{T}_u; \mb{\Theta}) &= - \sum_{u_j \in \mathcal{T}_u}  \sum_{k=1}^{|\mathcal{Y}|}\hat{\mb{y}}_{u,j}(k) \log {\mb{y}}_{u,j}(k),
\end{align}
\begin{align}
\ell(\mathcal{T}_s; \mb{\Theta}) &= - \sum_{s_l \in \mathcal{T}_s}  \sum_{k=1}^{|\mathcal{Y}|}\hat{\mb{y}}_{s,l}(k) \log {\mb{y}}_{s,l}(k),
\end{align}
where $\mb{y}_{u,j}$ and $\hat{\mb{y}}_{u,j}$ (and $\mb{y}_{s,l}$ and $\hat{\mb{y}}_{s,l}$) denote the prediction result vector and ground-truth vector of creator (and subject) respectively.

Formally, the main objective function of the {\dif} model can be represented as follows:
\begin{equation}
\min_{\mb{\Theta}} \ell(\mathcal{T}_n; \mb{\Theta}) + \ell(\mathcal{T}_u; \mb{\Theta}) + \ell(\mathcal{T}_s; \mb{\Theta}) + \alpha \cdot reg(\mb{\Theta}),
\end{equation}
where $\mb{\Theta}$ denotes all the involved variables to be learned, term ${reg}(\mb{\Theta})$ represents the regularization term (i.e., the sum of $L_2$ norm on the variable vectors and matrices), and $\alpha$ denotes the regularization term weight. By resolving the optimization functions, variables in the model can be effectively learned with the \textit{back-propagation} algorithm. For the news articles, creators and subjects in the testing set, their predicted labels will be outputted as the final result.



\subsection{{\gnl}: Graph Neural Lasso}

Graph neural lasso ({\gnl}) initially proposed in \cite{gnl} is a graph neural regression model and it can effectively incorporate the historical time-series data of multiple instances for addressing the dynamic network regression problem. {\gnl} extends the {\gdu} neuron \cite{dif} (also introduced in Section~\ref{subsec:gdu}) for incorporating both the network internal relationships and the network dynamic relationships between sequential network snapshots.

\subsubsection{Dynamic Gated Diffusive Unit}

{\gnl} also adopts {\gdu} as the basic neuron unit and extends it to the dynamic network regression problem settings, which can model both the network snapshot internal connections and the temporal dependency relationships between sequential network snapshots for the nodes. 


\begin{figure}[t]
    \centering
    \begin{minipage}{.5\textwidth}
    	\includegraphics[width=\linewidth]{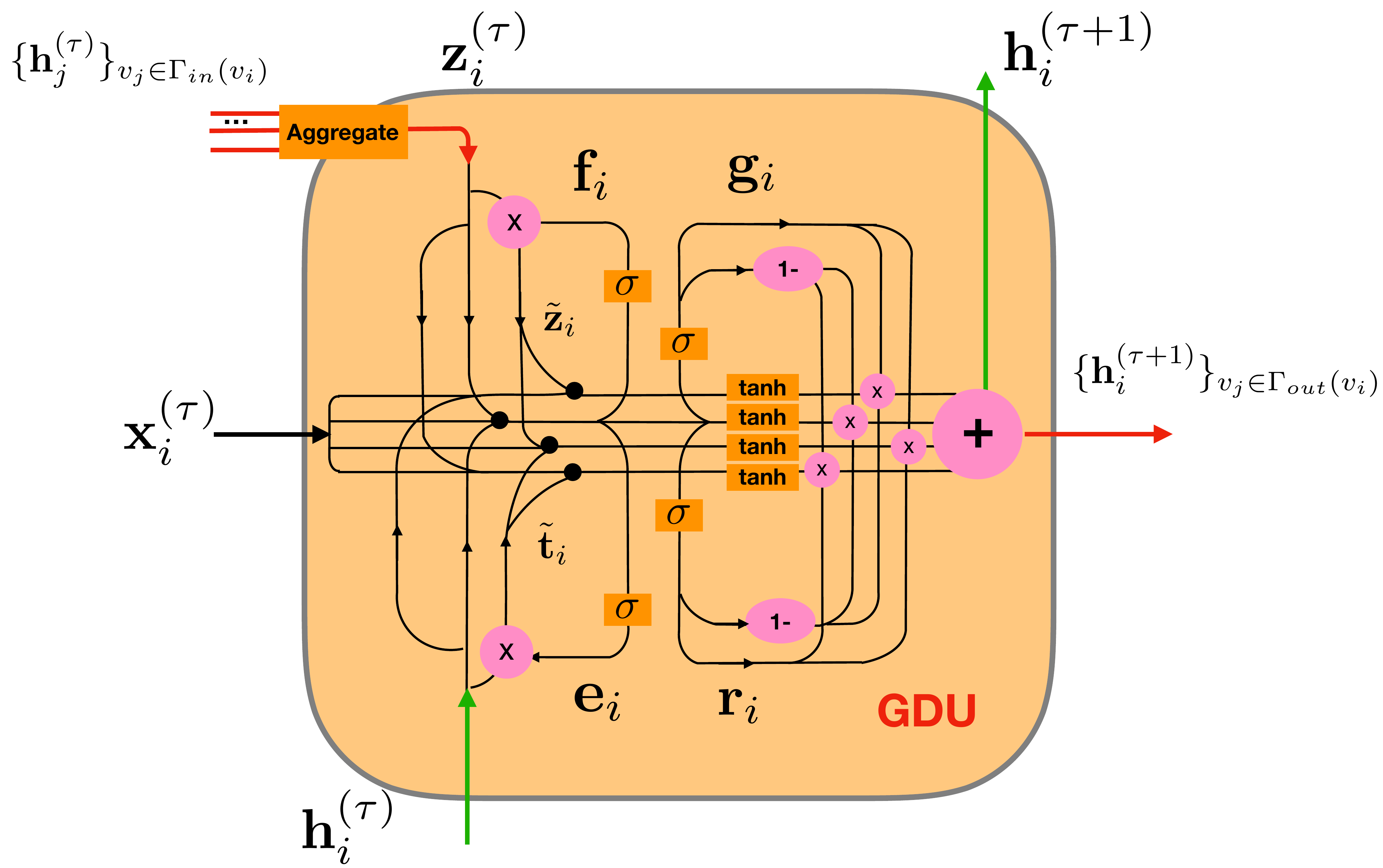}
    	\caption{The detailed architecture of the {\gdu} neuron of node $v_i$ at timestamp $\tau$.}
    	\label{fig:unit}
    \end{minipage}%
    \hfill
    \begin{minipage}{.42\textwidth}
    	\includegraphics[width=\linewidth]{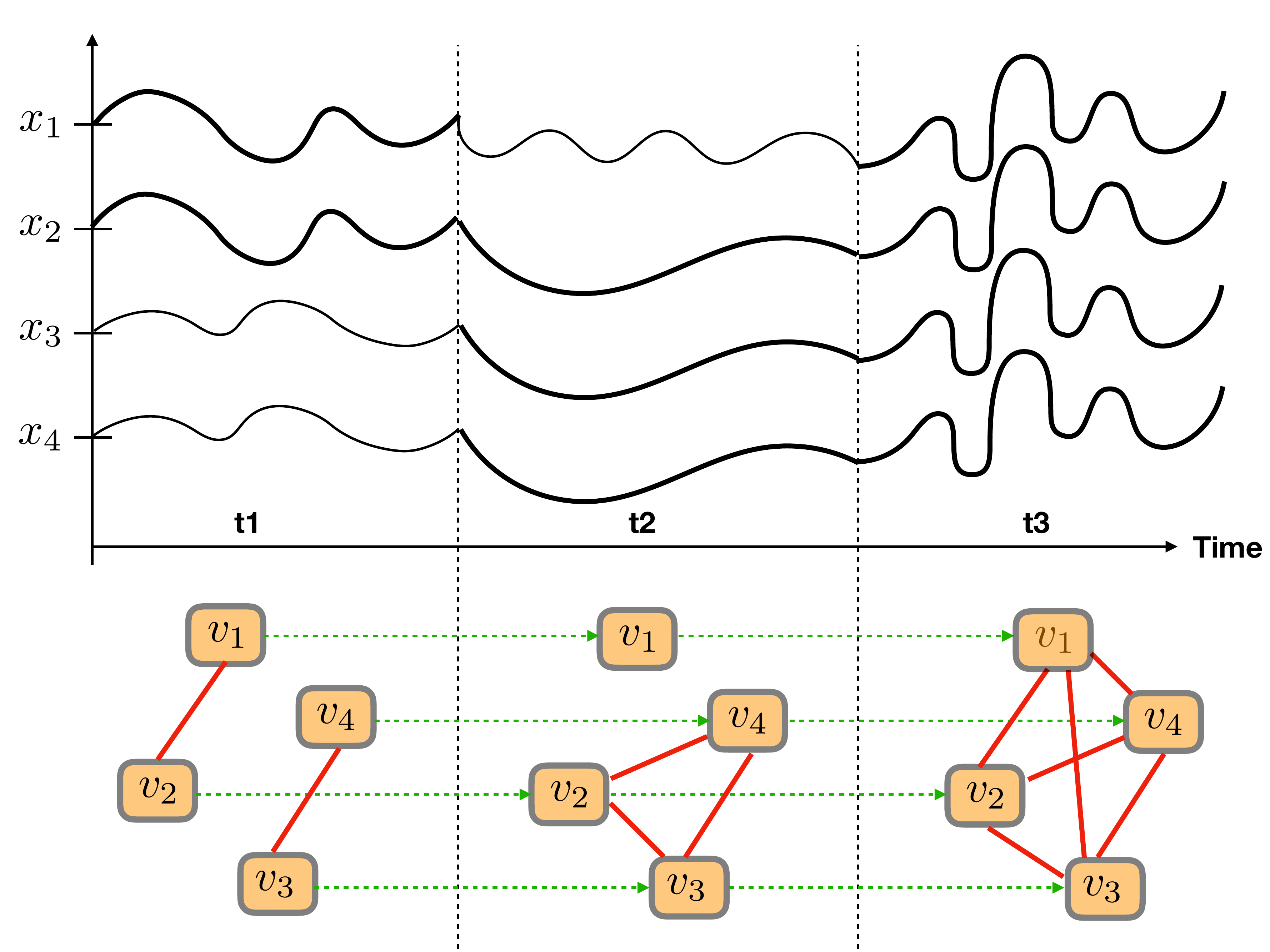}
    	\caption{The framework outline of {\gnl} based on {\gdu}.}
    	\label{fig:framework}
    \end{minipage}%

\end{figure}

Formally, given the time series data about a set of connected entities, such data can be represented as a dynamic network set $\mc{G} = \{G^{(1)}, G^{(2)}, \cdots, G^{(t)}\}$, where $t$ denotes the maximum timestamp. For each network $G^{(\tau)} \in \mc{G}$, it can be denoted as $G^{(\tau)} = (\mc{V}^{(\tau)}, \mc{E}^{(\tau)})$ involving the node set $\mc{V}^{(\tau)}$ and link set $\mc{E}^{(\tau)}$, respectively. Given a node $v_i$ in network $G^{(\tau)}$, its in-neighbors and out-neighbors in the network can be denoted as sets $\Gamma_{in}(v_i) = \{v_j | v_j \in \mc{V}^{(\tau)} \land (v_j, v_i ) \in \mc{E}^{(\tau)}\}$ and $\Gamma_{out}(v_i) = \{v_j | v_j \in \mc{V}^{(\tau)} \land (v_i, v_j ) \in \mc{E}^{(\tau)}\}$. Here, the link direction denotes the influences among the nodes. If the influences in the studied networks are bi-directional, the in/out neighbor sets will be identical, i.e., $\Gamma_{in}(v_i) = \Gamma_{out}(v_i)$.

For node $v_i$ in network $G^{(\tau)}$ of the $\tau_{th}$ timestamp, the input attribute values of $v_i$ can be denoted as an input feature vector $\mb{x}_i^{(\tau)} \in \mathbbm{R}^{d_x}$. {\gdu} maintains a hidden state vector for each node, and the vector of node $v_i$ at timestamp $\tau$ can be denoted as $\mb{h}_i^{(\tau)} \in \mathbbm{R}^{d_h}$. As illustrated in Figure~\ref{fig:unit}, besides the feature vector $\mb{x}_i^{(\tau)}$ and hidden state vector $\mb{h}_i^{(\tau)}$ inputs, the {\gdu} neuron of $v_i$ will also accept the inputs from $v_i$'s input neighbor nodes, i.e., $\{\mb{h}_j^{(\tau)}\}_{v_j \in \Gamma_{in}(v_i)}$, which will be integrated via certain aggregation operators:
\begin{equation}\label{equ:aggregate}
\mb{z}_i^{(\tau)} = \mbox{Aggregate} \left(\{\mb{h}_j^{(\tau)}\}_{v_j \in \Gamma_{in}(v_i)} \right).
\end{equation}
The $\mbox{Aggregate}(\cdot)$ operator used in {\gnl} will be introduced in detail in the next subsection.

A common problem with existing graph neural network models is over-smoothing, which will reduce all the nodes in the network to similar hidden representations. Such a problem will be much more serious when the model involves a deep architecture with multiple layers. To resolve such a problem, besides the attention mechanism to be introduced later, {\gdu} introduces several gates for the neural state adjustment as introduced in Section~\ref{subsec:gdu}. Formally, based on the input vectors $\mb{x}_i^{(\tau)}$, $\mb{z}_i^{(\tau)}$ and $\mb{h}_i^{(\tau)}$, the representation of node $v_i$ in the next timestamp $\tau + 1$ can be represented as
\begin{equation}\label{equ:dynamic_gdu}
\mb{h}_i^{(\tau + 1)} = {\gdu} \left( \mb{x}_i^{(\tau)}, \mb{z}_i^{(\tau)}, \mb{h}_i^{(\tau)} ; \mb{\Theta} \right).
\end{equation}
The concrete representation of the $GDU(\cdot)$ function is similar to Equ.~(\ref{equ:gdu_gate1})-(\ref{equ:gdu_update}) introduced before, and $\mb{\Theta}$ denotes the variables involved in the {\gdu} neuron.

\subsubsection{Attentive Neighborhood Influence Aggregation}

In this part, we will introduce the $\mbox{Aggregate}(\cdot)$ operator used in Equ.~(\ref{equ:aggregate}) for node neighborhood influence integration proposed in \cite{gnl}. The {\gnl} model defines such an operator based on an attention mechanism. Formally, given the node $v_i$ and its in-neighbor set $\Gamma_{in}(v_i)$, for any node $v_j \in \Gamma_{in}(v_i)$, {\gnl} quantifies the influence coefficient of $v_j$ on $v_i$ based on their hidden state vectors $\mb{h}_j^{(\tau)}$ and $\mb{h}_i^{(\tau)}$ as follows: \begingroup\makeatletter\def\f@size{9.5}\check@mathfonts
\begin{equation}
\alpha_{j, i}^{(\tau)} = \mbox{AttInf}(e_{j,i}^{(\tau)}) = \frac{\exp(e_{j,i}^{(\tau)})}{\sum_{v_k \in \Gamma_{out}(v_j)} \exp(e^{(\tau)}_{j,k})}, 
\mbox{ where }
e_{j,i}^{(\tau)} = \mbox{Linear}( \mb{W}_a \mb{h}_j^{(\tau)} \sqcup \mb{W}_a \mb{h}_i^{(\tau)}; \mb{w}_a ).
\end{equation}\endgroup
In the above equation, operator $\mbox{Linear}(\cdot; \mb{w}_a)$ denotes a linear sum of the input vector parameterized by weight vector $\mb{w}_a$. According to \cite{gat}, out of the model learning concerns, the above influence coefficient term can be slightly changed by adding the LeakyReLU function into its definition. Formally, the final influence coefficient used in {\gnl} is represented as follows:\begingroup\makeatletter\def\f@size{8.0}\check@mathfonts
\begin{equation}
\alpha^{(\tau)}_{j, i}  =   \mbox{\small AttInf}(\mb{h}_j^{(\tau)}, \mb{h}_i^{(\tau)}; \mb{W}_a, \mb{w}_a) =  \frac{\exp ( \mbox{\small LeakyReLU} ( \mbox{\small Linear}( \mb{W}_a \mb{h}_j^{(\tau)} \sqcup \mb{W}_a \mb{h}_i^{(\tau)}; \mb{w}_a ) ) )}{\sum_{v_k \in \Gamma_{out}(v_j)} \exp ( \mbox{\small LeakyReLU} ( \mbox{\small Linear}( \mb{W}_a \mb{h}_j^{(\tau)} \sqcup \mb{W}_a \mb{h}_k^{(\tau)}; \mb{w}_a ) ) )}.
\end{equation}\endgroup

Considering that in our problem setting the links in the dynamic networks are unknown and to be inferred, the above influence coefficient term $\alpha_{j, i}$ actually quantifies the existence probability of the influence link $(v_j, v_i)$, i.e., the inference results of the links. Furthermore, based on the influence coefficient, the concrete representation of Equ.~(\ref{equ:aggregate}) will be represented as follows:
\begin{equation}
\mb{z}_i^{(\tau)} = \mbox{Aggregate} \left(\{\mb{h}_j^{(\tau)}\}_{v_j \in \Gamma_{in}(v_i)} \right) = \sigma \left( \sum_{v_j \in \Gamma_{in}(v_i)} \alpha^{(\tau)}_{j,i} \mb{W}_a \mb{h}_j^{(\tau)} \right).
\end{equation}

\subsubsection{Graph Neural Lasso Model Learning}

In this part, we will introduce the architecture of the {\gnl} model together with its learning settings. Formally, given the input dynamic network set $\mc{G} = \{G^{(1)}, G^{(2)}, \cdots, G^{(t)}\}$, {\gnl} shifts a window of size $\tau$ along the networks in the order of their timestamps. The network snapshots covered by the window, e.g., $G^{(k)}$, $G^{(k+1)}$, $\cdots$, $G^{(k+\tau-1)}$, will be taken as the model input of {\gnl} to infer the network $G^{(k+\tau)}$ in following timestamp (where $k, k+1, \cdots, k+\tau \in \{1, 2, \cdots, t\}$). According to the above descriptions, the inferred attribute values of all the nodes and their potential influence links in network $G^{(k+\tau)}$ can be represented as
\begin{equation}
\begin{aligned}
&\hat{\mb{x}}_i^{(\tau+1)} = \mbox{FC}( \mb{h}_{i}^{(\tau+1)}; \boldsymbol{\Theta} ), \forall v_i \in \mc{V}^{(\tau+1)}; \\
&\alpha^{(\tau)}_{j,i} = \mbox{AttInf}(\mb{h}_{j}^{(\tau+1)}, \mb{h}_{i}^{(\tau+1)}; \boldsymbol{\Theta}), \forall v_i, v_j \in \mc{V}^{(\tau+1)},
\end{aligned}
\end{equation}
In the above equation, term $\mb{h}_{i}^{(\tau+1)}$ is defined in Equ.~(\ref{equ:dynamic_gdu}) and $\boldsymbol{\Theta}$ covers all the involved variables used in the {\gnl} model. By comparing the inferred node attribute values, e.g., $\hat{\mb{x}}_i^{(\tau+1)}$, against the ground truth values, e.g., ${\mb{x}}_i^{(\tau+1)}$, the quality of the inference results by {\gnl} can be effectively measured with some loss functions, e.g., mean square error:
\begin{equation}
\ell(\boldsymbol{\Theta}) = \frac{1}{|\mc{V}^{(\tau+1)}|} \sum_{v_i \in \mc{V}^{(\tau+1)}} \ell(v_i; \boldsymbol{\Theta}) = \frac{1}{|\mc{V}^{(\tau+1)}|} \sum_{v_i \in \mc{V}^{(\tau+1)}} \left\| \hat{\mb{x}}_i^{(\tau+1)} - {\mb{x}}_i^{(\tau+1)}\right\|_2^2.
\end{equation}
In addition, similar to {\lasso}, to avoid overfitting, {\gnl} proposes to add a regularization term in the objective function to maintain the sparsity of the variables. Formally, the final objective function of the {\gnl} model can be represented as follows:
\begin{equation}
\min_{\boldsymbol{\Theta}} \ell(\boldsymbol{\Theta}) + \beta \cdot \left\| \boldsymbol{\Theta} \right\|_1,
\end{equation}
where term $\left\| \boldsymbol{\Theta} \right\|_1$ denotes the sum of the $L_1$-norm regularizer of all the involved variables in the model and $\beta$ is the hyper-parameter weight of the regularization term. More information about the model learning as well as how to handle the non-derivable $L_1$-norm regularization term is available in \cite{gnl}.


\subsection{{\sage}: Graph Sample and Aggregate}

{\sage} introduced in \cite{sage} is an inductive model which focuses on leveraging node feature information for effective network node embedding. Instead of training individual embedding for each node, {\sage} learns a function that generate embeddings by sampling and aggregating features from nearby neighbors.

\subsubsection{Framework Descriptions}

\begin{algorithm}[t]
\small
\caption{Algorithm {\sage}}
\label{alg:graph_neural_network_sage}
\begin{algorithmic}[1]
	\REQUIRE Network $G = (\mc{V}, \mc{E})$; Input Node Feature: $\{\mb{x}_i\}_{v_i \in \mc{V}}$; Model Depth: $K$.
\ENSURE  Learned Representations $\{\mb{z}_i\}_{v_i \in \mc{V}}$.
\STATE	{Initialize $\mb{h}_i^{(0)} = \mb{x}_i, \forall v_i \in \mc{V}$}
\FOR	{$k \in \{1, 2, \cdots, K\}$}
\FOR	{$v_i \in \mc{V}$}
\STATE	{$\mc{N}(v_i) = \mbox{Sample} \left(\Gamma(v_i) \right)$}
\STATE	{$\mb{h}_{\Gamma(v_i)}^{(k)} = \mbox{Aggregate}\left( \left\{\mb{h}_{j}^{(k-1)} | v_j \in \mc{N}(v_i) \right\} \right)$}
\STATE	{$\mb{h}_{i}^{(k)} = \sigma\left( \mb{W}^{(k)} \left(\mb{h}_{\Gamma(v_i)}^{(k)} \sqcup \mb{h}_i^{(k-1)} \right) \right)$}
\ENDFOR
\STATE	{$\mb{h}_{i}^{(k)} = \mbox{Normalize} \left(\mb{h}_{i}^{(k)} \right)$}
\ENDFOR
\STATE	{$\mb{z}_i = \mb{h}_{i}^{(K)}, \forall v_i \in \mc{V}$}
\end{algorithmic}
\end{algorithm}

As illustrated in Algorithm~\ref{alg:graph_neural_network_sage}, the {\sage} algorithm accepts the network structure $G = (\mc{V}, \mc{E})$, input raw feature vectors $\{\mb{x}_i\}_{v_i \in \mc{V}}$ and model depth $K$ as the inputs, which will return the learned representations of nodes in the network as the output. Several functions will be called in the algorithm, including \textit{Sample}($\cdot$), \textit{Aggregate}($\cdot$) and \textit{Normalize}($\cdot$).

The forward computational process of {\sage} works iteratively layers by layers and nodes by nodes. Formally, the representations of nodes at each layer can be represented as vectors $\{\mb{h}_i^{(k)}\}_{v_i \in \mc{V}, k \in \{1, 2, \cdots, K\}}$, where $\mb{h}_i^{(k)}$ denotes the representation of $v_i$ at the $k_{th}$ layer. For all the nodes, their representations at layer $0$ are initialized with their raw features, i.e., $\mb{h}_i^{(0)} = \mb{x}_i, \forall v_i \in \mc{V}$. Given a node $v_i$, its neighbors can be represented as set $\Gamma(v_i) = \{v_j | v_j \in \mc{V} \land (v_i, v_j) \in \mc{E}\}$. {\sage} calls an aggregation function to effectively aggregate the neighbors' representations. However, instead of directly working on the complete neighbor set $\Gamma(v_i)$, {\sage} proposes to sample a subset of the neighbors prior to information aggregation, which is denoted as 
\begin{equation}
\mc{N}(v_i) = \mbox{Sample}\left( \Gamma(v_i) \right)
\end{equation} 
where $\mc{N}(v_i) \subset \Gamma(v_i)$ has a fixed size for all the nodes in the network and the sampling process follows a uniform distribution.

At the $k_{th}$ layer, via aggregating all the representations of the nodes in $\mc{N}(v_i)$ from the $(k-1)_{th}$ layer, {\sage} defines the a pseudo-representation for $v_i$ as follows:
\begin{equation}\label{equ:sage_aggregate}
\mb{h}_{\Gamma(v_i)}^{(k)} = \mbox{Aggregate}\left( \left\{\mb{h}_{j}^{(k-1)} | v_j \in \mc{N}(v_i) \right\} \right).
\end{equation}
The concrete representations of the $\mbox{Aggregate}\left(\cdot\right)$ function will be introduced later.

By concatenating the computed pseudo-representation $\mb{h}_{\Gamma(v_i)}^{(k)}$ and its representation at the $(k-1)_{th}$ layer, {\sage} defines the node representation updating equation as follows:
\begin{align}\label{equ:sage_update}
\mb{h}_{i}^{(k)} &= \sigma\left( \mb{W}^{(k)} \left(\mb{h}_{\Gamma(v_i)}^{(k)} \sqcup \mb{h}_i^{(k-1)} \right) \right),\\
\mb{h}_{i}^{(k)} &= \mbox{Normalize} \left(\mb{h}_{i}^{(k)} \right) = \frac{\mb{h}_{i}^{(k)}}{ \left\| \mb{h}_{i}^{(k)} \right\|_2 },
\end{align}
where operator $\sqcup$ denotes the concatenation of two vectors and {\sage} normalizes the vector $\mb{h}_{i}^{(k)}$ by dividing it with its modulus.

\subsubsection{Aggregator Function}

The ``orderless'' property of the the neighbor nodes poses more challenges on the aggregation operator to be used in {\sage}. Besides the aggregate function used above, several other aggregators can also be used for the information integration, which are provided as follows:
\begin{itemize}
\item \textit{Mean Aggregator}: The {mean aggregator} is very similar to the propagation rules used in {\gcn} introduced in Section~\ref{subsec:gcn}, which replaces Equ.~(\ref{equ:sage_aggregate}) and Equ.~(\ref{equ:sage_update}) with the following equation instead:
\begin{equation}
\mb{h}_{i}^{(k)} = \sigma \left(\mb{W}^{(k)} \mbox{Mean} \left( \left\{\mb{h}_{i}^{(k-1)} \right\} \cup \left\{\mb{h}_{j}^{(k-1)} | v_j \in \mc{N}(v_i) \right\} \right) \right).
\end{equation}

\item \textit{LSTM Aggregator}: Much more complex aggregators, e.g., LSTM, can also be adopted for the nodes representation aggregation and updating in {\sage}. By randomly permuting the neighbors in set $\Gamma(v_i)$, LSTM can be applied on the unordered set, where the output of the last unit can be obtained as the output. In other words, Equ.~(\ref{equ:sage_aggregate}) can be updated as follows: 
\begin{equation}
\mb{h}_{\Gamma(v_i)}^{(k)} = \mbox{LSTM}\left( \left\{\mb{h}_{j}^{(k-1)} | v_j \in \mc{N}(v_i) \right\} \right)
\end{equation}

\item \textit{Pooling Aggregator}: The pooling aggregator works with a max pooling layer to integrate the information from the neighbors, and Equ.~(\ref{equ:sage_aggregate}) can be replaced as follows:
\begin{equation}
\mb{h}_{\Gamma(v_i)}^{(k)} = \mbox{max}\left( \left\{ \sigma \left( \mb{W}^{(k)} \mb{h}_{j}^{(k-1)} + \mb{b}^{(k)} \right)  | v_j \in \mc{N}(v_i) \right\} \right)
\end{equation}

\end{itemize}


\subsection{{\segen}: Sample and Ensemble Genetic Evolutionary Network}

\begin{figure*}
	\centering
	\includegraphics[width=1.0\textwidth]{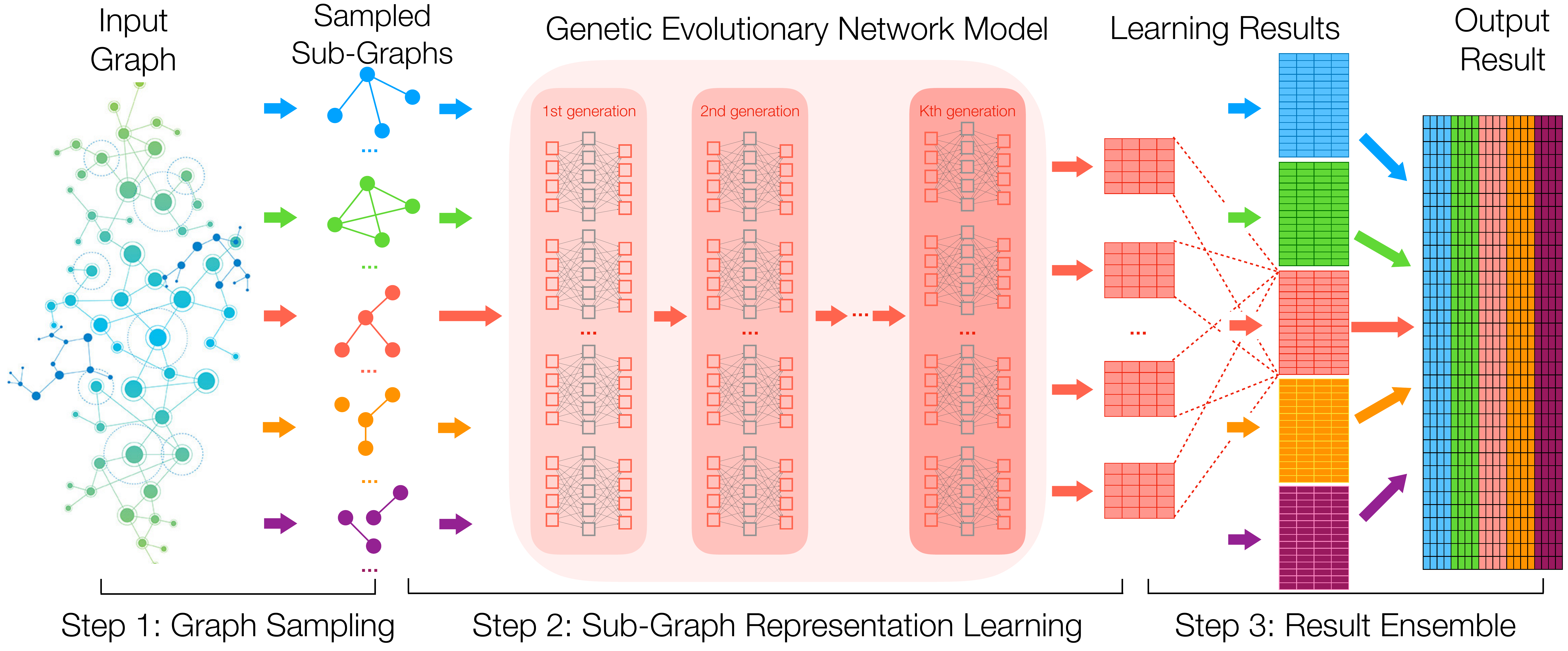}
	\caption{The {\segen} Framework \cite{segen} with Three Main Steps (Step 1: graph sampling to achieve a set of sub-graph instances; Step 2: sub-graph representation learning to get the representation features of nodes; Step 3: result ensemble of the learned node representations on each sub-graph by the unit models to get the final node representation).}
	\label{fig:framework}
	\vspace{-10pt}
\end{figure*}

Sample-Ensemble Genetic Evolutionary Network ({\segen}) first proposed in \cite{segen} can serve as an alternative approach to GNN models on giant networks. Instead of building one single graph neural network, based on a set of sampled sub- graphs, {\segen} adopts a genetic-evolutionary learning strategy to build a group of unit models generations by generations. The unit models incorporated in {\segen} can be either traditional graph representation learning models or the recent graph neural network models with a much ``narrower'' and ``shallower'' architecture. The learning results of each instance at the final generation will be effectively combined from each unit model via diffusive propagation and ensemble learning strategies.

\subsubsection{{\segen} Architecture Description}

In framework {\segen}, instead of handling the input large-scale graph data directly, it proposes to sample a subset (of set size $s$) of small-sized sub-graphs (of a pre-specified sub-graph size $k$) instead and learn the representation feature vectors of nodes based on the sub-graphs. To ensure the learned representations can effectively represent the characteristics of nodes, {\segen} need to ensure the sampled sub-graphs share similar properties as the original large-sized input graph. As shown in Figure~\ref{fig:framework}, five different types of graph sampling strategies (indicated in five different colors) are adopted, and each strategy will lead to a group of small-sized sub-graphs, which can capture both the local and global structures of the original graph. According to \cite{segen}, the sampled sub-graphs based on different sampling strategies, e.g., BFS, DFS, BNS, BES and HS, can be represented as $\mathcal{G}^{\textsc{bfs}}$, $\mathcal{G}^{\textsc{dfs}}$, $\mathcal{G}^{\textsc{hs}}$, $\mathcal{G}^{\textsc{ns}}$ or $\mathcal{G}^{\textsc{es}}$, respectively.

Instead of fitting each unit model with all the sub-graphs in the pool $\mathcal{G}$, in the unit model, a set of sub-graph training batches $\mathcal{T}_1, \mathcal{T}_2, \cdots, \mathcal{T}_m$ will be sampled for each unit model respectively in the learning process, where $|\mathcal{T}_i| = b, \forall i \in \{1, 2, \cdots, m\}$ are of the pre-defined batch size $b$. These batches may share common sub-graphs as well, i.e., $\mathcal{T}_i \cap \mathcal{T}_j$ may not necessary be $\emptyset$. In the model, the unit models learning process for each generation involves two steps: (1) generating the batches $\mathcal{T}_i$ from the pool set $\mathcal{G}$ for each unit model $M^1_i \in \mathcal{M}^1$, and (2) learning the variables of the unit model $M^1_i$ based on sub-graphs in batch $\mathcal{T}_i$. Considering that the unit models have a much smaller number of hidden layers, the learning time cost of each unit model will be much less than the deeper models on larger-sized graphs.

In the following parts, we will first introduce the learning process of the model, which accepts each sub-graph pool as the input and learns the representation feature vectors of nodes as the output. We can use $\mathcal{G}$ to represent the sampled pool set, which can be $\mathcal{G}^{\textsc{bfs}}$, $\mathcal{G}^{\textsc{dfs}}$, $\mathcal{G}^{\textsc{hs}}$, $\mathcal{G}^{\textsc{ns}}$ or $\mathcal{G}^{\textsc{es}}$ respectively. The learned results can be further fused together with a hierarchical ensemble process to be introduced at the last subsection.

\subsubsection{Genetic Evolutionary Learning of {\segen}}

The training process of {\segen} involves several key steps, including \textit{unit model evaluation and selection}, \textit{crossover } and \textit{mutation}, which will be introduced as follows:
\begin{itemize}

\item \textbf{Evaluation and Selection}: The unit models in the generation set $\mathcal{M}^1$ can have different performance, due to (1) different initial variable values, and (2) different training batches in the learning process. In framework {\segen}, instead of applying ``deep'' models with multiple hidden layers, it proposes to ``deepen'' the models in another way: ``evolve the unit model into `deeper' generations''. For each unit model $M^1_k \in \mathcal{M}^1$, based on the sub-graphs in a validation set $\mathcal{V}$, {\segen} measures the introduced loss of the model as
\begin{align*}
\ell(M^1_k; \mathcal{V}) = \sum_{g \in \mathcal{V}} \sum_{v_i, v_j \in \mathcal{V}_g, v_i \neq v_j} s_{i,j} \left\| \mb{z}_{k,i}^1 - \mb{z}_{k,j}^1 \right\|_2^2,
\end{align*}
where $\mb{z}_{k,i}^1$ and $\mb{z}_{k,j}^1$ denote the learned latent representation feature vectors of nodes $v_i, v_j$ in the sampled sub-graph $g$. Term $s_{i,j}$ has value $+1$ if $v_i$ and $v_j$ are connected in subgraph $g$, otherwise $s_{i,j}$ will be assigned with value $0$ instead.

The probability for each unit model to be picked as the parent model for the \textit{crossover} and \textit{mutation} operations can be represented as
\begin{align*}
p(M^1_k) = \frac{\exp^{- \ell(M^1_k; \mathcal{V})}}{ \sum_{M^1_i \in \mathcal{M}^1} \exp^{-\ell(M^1_i; \mathcal{V})}}.
\end{align*}
In the real-world applications, a normalization of the loss terms among these unit models is necessary. For the unit model introducing a smaller loss, it will have a larger chance to be selected as the parent unit model. Considering that the \textit{crossover} is usually done based a pair of parent models, the pairs of parent models selected from set $\mathcal{M}^1$ can be denoted as $\mathcal{P}^1 = \{(M^1_i, M^1_j)_k\}_{k \in \{1, 2, \cdots, m\}}$, based on which {\segen} will be able to generate the next generation of unit models, i.e., $\mathcal{M}^2$.

\item \textbf{Crossover}: For the $k_{th}$ pair of parent unit model $(M^1_i, M^1_j)_k \in \mathcal{P}^1$, their genes can be denoted as their variables $\mb{\theta}^1_i, \mb{\theta}^1_j$ respectively (since the differences among the unit models mainly lie in their variables), which are actually their chromosomes for crossover and mutation.

{\segen} proposes to adopt the \textit{uniform crossover} to get the chromosomes (i.e., the variables) of their child model. Considering that the parent models $M^1_i$ and $M^1_j$ can actually achieve different performance on the validation set $\mathcal{V}$, in the crossover, the unit model achieving better performance should have a larger chance to pass its chromosomes to the child model. 

Formally, the chromosome inheritance probability for parent model $M^1_i$ can be represented as
\begin{alignat*}{2}
p(M^1_i) = \frac{\exp^{- \ell(M^1_i; \mathcal{V})}}{\exp^{-\ell(M^1_i; \mathcal{V})} + \exp^{-\ell(M^1_j; \mathcal{V})}}
\end{alignat*}
Meanwhile, the chromosome inheritance probability for model $M^1_j$ can be denoted as $p(M^1_j) = 1- p(M^1_i)$.

In the uniform crossover method, based on parent model pair $(M^1_i, M^1_j)_k \in \mathcal{P}^1$, the obtained child model chromosome vector can be denoted as $\mb{\theta}^2_k \in \mathbb{R}^{ | \mb{\theta}^1 | }$ (the superscript denotes the $2_{nd}$ generation and $| \mb{\theta}^1 |$ denotes the variable length), which is generated from the chromosome vectors $\mb{\theta}^1_i$ and $\mb{\theta}^1_j$ of the parent models. Meanwhile, the crossover choice at each position of the chromosomes vector can be represented as a vector $\mb{c} \in \{i, j\}^{| \mb{\theta}^1 |}$. The entries in vector $\mb{c}$ are randomly selected from values in $\{i, j\}$ with a probability $p(M^1_i)$ to pick value $i$ and a probability $p(M^1_j)$ to pick value $j$ respectively. The $l_{th}$ entry of vector $\mb{\theta}^2_k$ before mutation can be represented as
\begin{align*}
\hat{{\theta}}^2_k(l) = \mathbbm{1}\left(c(l) = i \right) \cdot {\theta}^1_i(l) + \mathbbm{1}\left(c(l) = j \right) \cdot {\theta}^1_j(l),
\end{align*}
where indicator function $\mathbbm{1}(\cdot)$ returns value $1$ if the condition is True; otherwise, it returns value $0$.

\item \textbf{Mutation}: The variables in the chromosome vector $\hat{{\theta}}^2_k(l) \in \mathbb{R}^{| \mb{\theta}^1 |}$ are all real values, and some of them can be altered, which is also called \textit{mutation} in traditional genetic algorithm. Mutation happens rarely, and the chromosome mutation probability is $\gamma$ in the model. Formally, the mutation indicator vector can be denoted as $\mb{m} \in \{0, 1\}^d$, and the $l_{th}$ entry of vector $\mb{\theta}^2_k$ after mutation can be represented as
\begin{align*}
{{\theta}}^2_k(l) = \mathbbm{1}\left(m(l) = 0 \right) \cdot \hat{{\theta}}^2_k(l) + \mathbbm{1}\left(c(l) = 1 \right) \cdot rand(0, 1),
\end{align*}
where $rand(0, 1)$ denotes a random value selected from range $[0, 1]$. Formally, the chromosome vector $\mb{\theta}^2_k$ defines a new unit model with knowledge inherited form the parent models, which can be denoted as $M^2_k$. Based on the parent model set $\mathcal{P}^1$, all these newly generated models can be represented as $\mathcal{M}^2 = \{M^2_k\}_{(M^1_i, M^1_j)_k \in \mathcal{P}^1}$, which will form the $2_{nd}$ generation of unit models.

\end{itemize} 

\subsubsection{Result Ensemble}

Based on the models introduced in the previous subsection, {\segen} adopts a hierarchical result ensemble method, which involves two steps: (1) \textit{local ensemble} of results for the sub-graphs on each sampling strategies, and (2) \textit{global ensemble} of results obtained across different sampling strategies.

\begin{itemize}
\item \textbf{Local Ensemble}: Based on the sub-graph pool $\mathcal{G}$ obtained via the sampling strategies introduced before, we have learned the $K_{th}$ generation of the unit model $\mathcal{M}^K$ (or $\mathcal{M}$ for simplicity), which contains $m$ unit models. Formally, given a sub-graph $g \in \mathcal{G}$ with node set $\mathcal{V}_{g}$, by applying unit model $M_j \in \mathcal{M}$ to $g$, the learned representation for node $v_q \in \mathcal{V}_{g}$ can be denoted as vector $\mb{z}_{j,q}$, where $q$ denotes the unique node index in the original complete graph $G$ before sampling. For the nodes $v_p \notin \mathcal{V}_{g}$, its representation vector will be $\mb{z}_{j,p} = \mb{null}$, which denotes a dummy vector of length $d$. Formally, the learned representation feature vector for node $v_q$ can be represented as
\begin{equation}
\mb{z}_q = \bigsqcup_{g \in \mathcal{G}, M_j \in \mathcal{M}, } \mb{z}_{j,q},
\end{equation}
where operator $\sqcup$ denotes the concatenation operation of feature vectors. Considering that in the graph sampling step, not all nodes will be selected in sub-graphs. For the nodes $v_p \notin \mathcal{V}_{g}, \forall g \in \mathcal{G}$, {\segen} introduces a local propagation approach to compute its representations based on its neighbors instead.

\textbf{Global Ensemble}: Generally, these different graph sampling strategies can capture different local/global structures of the graph, which will all be useful for the node representation learning. In the global result ensemble step, {\segen} proposes to group these features together as the output. Formally, for node $v_q$ in the original network, its fused representations can be denoted as
\begin{equation}
\bar{\mb{z}}_q = \sum_{i \in \{\textsc{bfs}, \textsc{dfs}, \textsc{hs}, \textsc{ns}, \textsc{es} \}} w^i \cdot \mb{z}_q^i,
\end{equation}
where $\mb{z}_q^{\textsc{bfs}}$, $\mb{z}_q^{\textsc{dfs}}$, $\mb{z}_q^{\textsc{hs}}$, $\mb{z}_q^{\textsc{ns}}$ and $\mb{z}_q^{\textsc{es}}$ are the vectors of $v_q$ obtained from the above local ensemble based on different graph sampling strategies. In \cite{segen}, {\segen} will simply assign them with equal weights, i.e., $\bar{\mb{z}}_q $ is an average of $\mb{z}_q^{\textsc{bfs}}$, $\mb{z}_q^{\textsc{dfs}}$, $\mb{z}_q^{\textsc{hs}}$, $\mb{z}_q^{\textsc{ns}}$ and $\mb{z}_q^{\textsc{es}}$.
\end{itemize}


\section{Challenges and Opportunities}

We have also observed many challenges with graph neural network studies, which provide plenty of opportunities for researchers interested in this topic:

\subsection{Over-Smoothing Problem}

The existing graph neural networks on giant network representation learning problem suffer from the \textit{over-smoothing problem} a lot. For instance, if a {\gcn} is deep with
many convolutional layers, the output features may be over-smoothed and vertices from different clusters may become indistinguishable, which will render the {\gcn} model fail to work. We have also observed some approaches proposed to resolve such a problem. In \cite{gnl, dif}, both the {\dif} and {\gnl} models introduce a set of gates (i.e., the {\gdu} neuron unit) to ensure the nodes can capture the raw inputs, neighbors' influences and the temporal dynamic states in the learning process, which can resolve the over-smoothing problem effectively.

\subsection{Optimization Efficiency}

The time complexity of the graph neural network learning, including those for small graphs and giant networks, can be very high. For the small graph oriented graph neural networks, e.g., {\isonn}, the major time is spent on enumerating the permutation matrices for the isomorphic feature calculation. Meanwhile, for the giant network oriented graph neural networks, e.g., {\gcn} and {\gat}, most of the time is spent on the backpropagation along the graph edges for the nodes, which grows almost quadratically as the network size increases and exponentially as the model goes deeper. New optimization algorithms that can address the high learning cost will be necessary and desired.

\subsection{The Gap Between Small Graphs and Giant Networks}

By this context so far, we haven't witnessed any graph neural networks that can learn effective representations for both the small graphs and the giant networks simultaneously without any architecture modifications. Proposing a new unified graph neural network model that can work for different types of networks will be desired.

\subsection{Graph Neural Network for Dynamic Networks}

Most of the network data in the real-world are not static, which keep changing with time. We have observed the {\gnl} model \cite{gnl} as the first work focusing on the dynamic regression scenario. Such kinds of model can be important, and it may also serve as the tool for analyzing and understanding the brain activities, which is a dynamic network with both structure and states changing all the time.

\subsection{Graph Neural Network for Complex Networks}

Most of the graph neural networks proposed so far mainly focus on the homogeneous network, which over-simplify the learning setting, since most of the network data in the real world are heterogeneous instead. Generally, in heterogeneous networks, different types of nodes and links can convey different kinds of physical meanings. New models that can effective incorporate such heterogeneous information in the learning process can be desired for concrete real-world applications of graph neural networks.

\section{Summary}

In this paper, we have introduced the latest graph neural networks proposed for resolving the small graph and giant network oriented research problems. The small graph oriented graph neural network models introduced in this paper include {\isonn}, {\sdbn} and {\lfer}; whereas, the giant network oriented graph neural network models introduced in this paper include {\gcn}, {\gat}, {\dif}, {\gnl}, {\sage} and {\segen}. In addition, we have also introduced several challenges and opportunities on graph neural network studies. This paper will be updated shortly as we observe the new development on this topic in the near future.



\newpage

\vskip 0.2in
\bibliographystyle{plain}
\bibliography{reference}

\end{document}